\documentclass{article}

\usepackage{microtype}
\usepackage{graphicx}
\usepackage{subcaption}
\usepackage{booktabs} 
\usepackage{subcaption}
\usepackage{hyperref}

\usepackage{configuration/conference,times}

\usepackage{amsmath}
\usepackage{amssymb}
\usepackage{mathtools}
\usepackage{amsthm}
\usepackage{multirow}
\usepackage{graphicx}

\usepackage[capitalize,noabbrev]{cleveref}

\usepackage[textsize=tiny]{todonotes}
\usepackage{amsmath,amssymb,amsthm}
\newtheorem{theorem}{Theorem}[section]

\newtheorem{definition}[theorem]{Definition}
\newtheorem{example}[theorem]{Example}

\newenvironment{talign*}
{\csname align*\endcsname}
{\endalign}

\usepackage[utf8]{inputenc}         %
\usepackage[T1]{fontenc}            %
\usepackage{url}                    %
\usepackage{booktabs}               %
\usepackage{amsfonts}               %
\usepackage{nicefrac}               %
\usepackage{microtype}              %
\usepackage{xcolor}                 %
\usepackage{algorithm}
\usepackage{algorithmic}
\usepackage{graphicx}
\usepackage{subcaption}
\usepackage[flushleft]{threeparttable}
\usepackage{float}
\usepackage{multirow}
\usepackage{xspace}
\usepackage{natbib}
\usepackage{enumitem}
\usepackage[font=small]{caption}
\usepackage{autobreak}
\usepackage{sidecap}
\usepackage{wrapfig}
\usepackage{bbding}
\usepackage[toc, page, header]{appendix}
\usepackage{tikz}
\usepackage{xcolor}
\usepackage{pifont}
\usepackage{mdframed}

\hypersetup{
    colorlinks=true,
    linkcolor=blue,
    citecolor=blue,
    urlcolor=blue
}

\definecolor{coral}{RGB}{255,127,80}
\definecolor{darkgreen}{RGB}{0,100,0}
\definecolor{darkyellow}{RGB}{204,153,0}
\definecolor{salmon}{RGB}{250,128,114}

\usepackage{xspace}

\newcommand{\triage}{\textsc{Triage}\,}
\newcommand{\method}{\textsc{Triage-PiEvo}\,}
\newcommand{\pievo}{\textsc{PiEvo}\,}
\newcommand{\revise}{\mathsf{revise}\,}
\newcommand{\retain}{\mathsf{retain}\,}

\usepackage{needspace}

\definecolor{takeawaybg}{RGB}{242,237,250}
\definecolor{takeawayframe}{RGB}{150,125,200}
\newmdenv[
  backgroundcolor=takeawaybg,
  linecolor=takeawayframe,
  topline=false, bottomline=false, rightline=false,
  linewidth=2pt,
  innerleftmargin=5pt, innerrightmargin=5pt,
  innertopmargin=5pt, innerbottommargin=3pt,
  skipabove=5pt, skipbelow=1pt
]{takeaway}

\definecolor{defbg}{RGB}{252,238,238}
\definecolor{defframe}{RGB}{196,98,98}
\newmdenv[
  backgroundcolor=defbg,
  linecolor=defframe,
  topline=false, bottomline=false, rightline=false,
  linewidth=2pt,
  innerleftmargin=5pt, innerrightmargin=5pt,
  innertopmargin=5pt, innerbottommargin=3pt,
  skipabove=5pt, skipbelow=1pt
]{defbox}

\usepackage{xurl}

\title{Triage-PiEvo: Evidence Admission for Principle-Evolving Scientific Agents}

\author{%
Peize Li$^{1,2,*}$ \quad Yingming Pu$^{1,5,*}$ \quad Xingguo Xu$^{3,2}$ \\[2pt]
\bf Yiming Cui$^{4,2}$ \quad Yuansheng Gao$^{5,2}$ \quad Tao Lin$^{1,\dagger}$ \mdseries \\[6pt]
$^{1}$Westlake University \quad $^{2}$Kuaishou Technology \quad $^{3}$Dalian University of Technology \\[2pt]
$^{4}$Hangzhou Dianzi University \quad $^{5}$Zhejiang University \\[6pt]
$^{*}$Equal contribution \quad $^{\dagger}$Corresponding author \\[2pt]
\ding{41}~\texttt{lipeize05@kuaishou.com}, \texttt{\{puyingming, lintao\}@westlake.edu.cn}
}

\finalcopy

\newcommand\blfootnote[1]{%
  \begingroup
  \renewcommand\thefootnote{}\footnote{#1}%
  \addtocounter{footnote}{-1}%
  \endgroup
}

\begin{document}

\maketitle
\lhead{}  
\blfootnote{Preprint.}

\begin{abstract}
Large language model (LLM)-based AI Scientists advance scientific discovery, yet they suffer from inefficiencies due to noisy observations.
Existing principle-evolving agents act on every anomaly, leaving its source unresolved: a new mechanism or an evaluation artifact, so invalid revisions bias evolution.
To address this, we formulate revision as \emph{evidence admission}: anomalies steer search immediately, but only independently verified evidence may rewrite persistent principles, and contradicted principles are retracted.
We realize this in \textbf{\method{}}, which pairs an anomaly-responsive search portfolio with a staged verification gate that alone controls what enters principle state.
Across four scientific tasks with three backbones, \method{} (1) attains the highest average solution quality with $+7.5\%\sim11.8\%$ over the strongest baseline per backbone, matching the strongest baseline's full-budget quality with one-third of the evaluation budget on Qwen3-235B; and additional stress tests and cross-harness deployment demonstrate that (2) in constructed stress tests, it admits none of the corrupted, hidden-source, or exploit evidence that ungated variants admit, while missing no planted mechanism, and (3) plugged unchanged into three external research harnesses, it cuts false evidence admissions by over $60\%$ and raises verified scores by $31\%\sim43\%$.
\end{abstract}

\section{Introduction}

\begin{wrapfigure}[15]{r}{0.35\textwidth}
    \vspace{-12pt}
    \centering
    \includegraphics[width=0.88\linewidth]{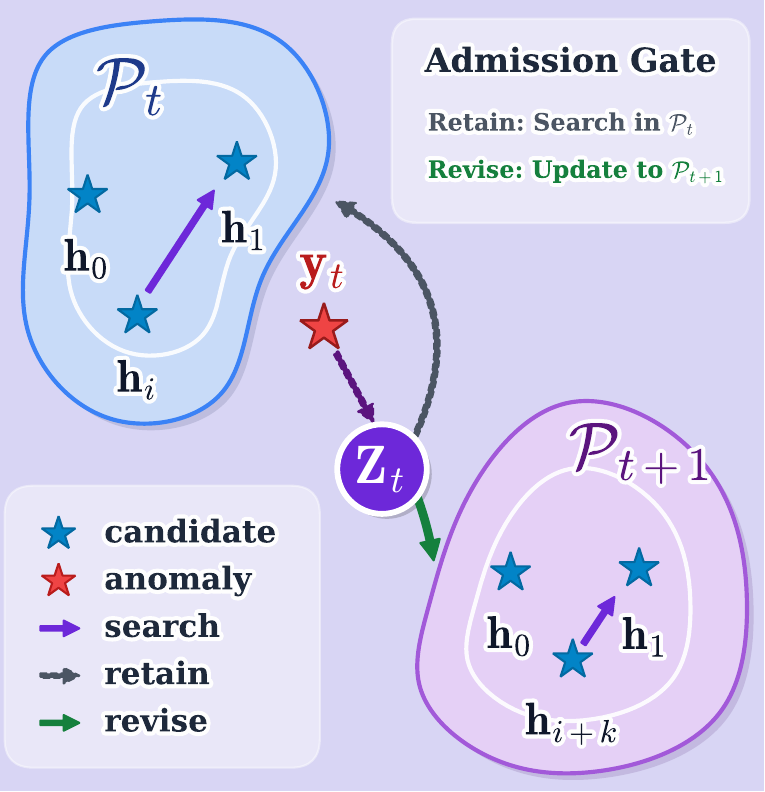}
    \vspace{-5pt}
    \caption{\textbf{Source-aware triage.} Latent source $Z_t$ mediates the verified retain-or-revise decision.}
    \label{fig:triage-overview}
    \vspace{-12pt}
\end{wrapfigure}

Autonomous scientific agents now automate scientific research within a single closed loop~\citep{lu2024ai,yamada2025ai,schmidgall2025agent,gottweis2025towards}. 
On the premise that true knowledge often lies outside the initial hypothesis space, principle-evolution agents~\citep{pu2026principle,pu2025piflow} make every anomaly a decision in this loop: steer hypothesis search, or rewrite the persisted theory and explanation.

While surprise-driven revision~\citep{pu2026principle} lets agents escape their initial hypothesis space, it treats the observation's reliability as given. Consequently, the revision may degrade because it rests on a signal that cannot discriminate, trading false revisions against missed theory blindly.
Nor do existing methods cover this gap: verification loops critique transient responses rather than persisted theory~\citep{shinn2023reflexion,madaan2023self,gou2024critic,dhuliawala2024chain}, and self-correction without external evidence is unreliable~\citep{huang2024large,kamoi2024can}. 
Accordingly, community evaluations find that auto-research agents produce extensive artifacts yet seldom revise their internal theory under contradictory evidence~\citep{rios2026ai,kirgis2026can}. 
This rigidity leads to significant inefficiencies: \emph{anomalies are read as evidence of science, corrupted feedback enters trusted scientific state}.

These limitations reduce to three challenges: (a) one anomaly score has \textbf{source ambiguity} that cannot tell a new mechanism from corrupted or misspecified evaluations; (b) the \textbf{role conflation} of one history serves both search and principle acceptance, so unreliable observations can rewrite trusted principles; and (c) the \textbf{revision irreversibility} that accepted principles cannot be retracted under contradictory evidence.

To bridge this gap, we propose \textbf{\triage}, an evidence-admission layer that decouples anomaly detection from principle revision~\citep{pu2026principle}. 
\triage treats each anomaly as source-ambiguous and routes evidence along two separately logged paths: (a) observations steer candidate selection, while (b) verification records update per-channel reliability estimates.
A principle revision is trusted only after it survives independent tests on unseen cases, as shown in Figure~\ref{fig:triage-overview}.
We instantiate this in \textbf{\method{}}, which pairs \triage{} with Active Flow Expansion~\citep{de2026active}: the portfolio allocates the evaluation budget across exploitation, frontier expansion, and mechanism-discriminating probes, while the evidence-admission decides which evidence alters principle state.

We evaluate \method on four discovery tasks under three backbone LLMs.
Empirical results demonstrate that \textbf{\method}: 
(a) \textbf{attains the highest performance}, raising average solution quality from \pievo's $62.87$--$70.35$\% to $79.63$--$83.37$\%; 
(b) \textbf{admits no corrupted or exploit-region observations} in constructed stress tests where gate-off variants admit every one, under evaluator corruption and proxy exploitation alike (the latter further raising held-out score by $0.209$);
(c) \textbf{cuts false expansions} on injected noise, surrogate failure, and non-generalizing local artifacts from rate of $1.0$ to $0.0$ in the same tests without missing any planted mechanism, and backtracks to the correct state after mechanism reversal; and
(d) \textbf{transfers unchanged to three external research harnesses}, cutting false evidence admissions by over $60\%$ and raising verified scores by $31\%\sim43\%$.

In summary, our contributions are:
\begin{enumerate}[nosep, leftmargin=16pt]
  \item[(a)] We introduce \textbf{Evidence Admission} as the decision layer of anomaly-driven agents for changing scientific state. Anomalies are treated as source-ambiguous, and revision is governed by asymmetric costs of false revision versus missed mechanism.
  \item[(b)] We propose \textbf{\method{}}, which realizes this layer through per-channel trust estimation, a principle lifecycle gated by independent held-out support.
  \item[(c)] We conduct \textbf{extensive evaluations} across four scientific tasks and three backbones: \method{} raises average solution quality by $+7.5\%\sim11.8\%$ over the strongest baseline per backbone, and controlled corruption, hidden-source, and proxy-exploitation studies show that the gate rejects all unreliable evidence injected in these studies while missing no planted mechanism.
\end{enumerate}

\section{Related Work}
\label{sec:related}

\noindent \textbf{Autonomous scientific agents and evolving state. }
Scientific agents automate growing shares of the research cycle, from end-to-end paper production and multi-stage pipelines~\citep{lu2024ai,yamada2025ai,tang2026ai,schmidgall2025agent,gottweis2025towards} to tool-coupled domain execution and closed-loop laboratory discovery~\citep{bran2023chemcrow,boiko2023autonomous,ruan2024automatic,yang2025moose,ghafarollahi2025sciagents,ding2025scitoolagent,ghareeb2026multi,narayanan2024aviary,laurent2024lab,mitchener2025bixbench,skarlinski2024language,mitchener2025kosmos}. What persists between iterations there is an artifact, not a revisable scientific account. A narrower line makes the internal state editable: \pievo{} and PiFlow expand a principle space on conflicting observations~\citep{pu2026principle,pu2025piflow}, ATLAS selects mechanism-discriminating experiments~\citep{eltetHo2026atlas}, self-revising frameworks log regime changes~\citep{wang2026self}, and evolving architectures such as EvoMaster~\citep{zhu2026evomaster} and ADMET-EvO~\citep{zhou2026admet} pursue cumulative adaptation across tasks, with agentic laboratory world models unifying hypothesis generation and state tracking~\citep{fu2026agentic}. Editable state is also corruptible: without explicit evidence admission, anomalous or corrupted feedback rewrites persistent principles directly~\citep{rios2026ai,kirgis2026can}. \triage{} separates the history that steers search from the evidence authorized to alter scientific state.

\noindent \textbf{Active experimentation under heterogeneous sources.}
In active learning and Bayesian optimization, one line decides where to spend evaluations~\citep{mackay1992information,houlsby2011bayesian}, constraining exploration for safety~\citep{sui2015safe} or scalability~\citep{eriksson2019scalable}. Another exploits heterogeneous costs, fidelities, and failure rates across sources~\citep{swersky2013multi,kandasamy2017multi,poloczek2017multi,ghoreishi2019multi}, extending to unreliable channels~\citep{mikkola2023multi,fan2024multi}, quality-controlled loops~\citep{harris2025active}, and out-of-distribution expansion~\citep{de2026active}. Both decide which observation to acquire \emph{before} it is run, so an anomalous outcome inherits whatever reliability the acquisition rule assumed. The difference is in what the decision protects: multi-source methods weight a noisy source inside a regression model, while the object at risk here is a persistent principle that a single admitted artifact can corrupt for the rest of the run. Our AFE portfolio takes the acquisition-side view for candidate generation; \triage{} makes the post-observation decision of whether an anomaly may update that principle.

\noindent \textbf{Verification, provenance, and agent monitoring.}
In agent systems, verification targets transient outputs through feedback recycling~\citep{yao2022react,shinn2023reflexion,madaan2023self}, tool-based checking~\citep{gou2024critic}, independent verification questions~\citep{dhuliawala2024chain}, and evidence accumulation~\citep{zhang2026guideline}, though self-correction without external evidence is unreliable~\citep{huang2024large,kamoi2024can}. Related work argues for monitoring agentic systems before trusting them~\citep{boston2026monitoring} and for immutable provenance trails over agent-generated artifacts~\citep{liu2026last}, while Sibyl~\citep{rangelov2026sibyl} uses temporal backtesting as held-out evidence. \triage{} carries these provenance and monitoring ideas to persistent principles: same-source agreement keeps observations search-active, acceptance requires independent held-out support, and contradictory held-out evidence forces retraction.

\section{Methodology}
\label{sec:methodology}
\begin{figure*}[t!]
    \centering
    \includegraphics[width=0.88\textwidth]{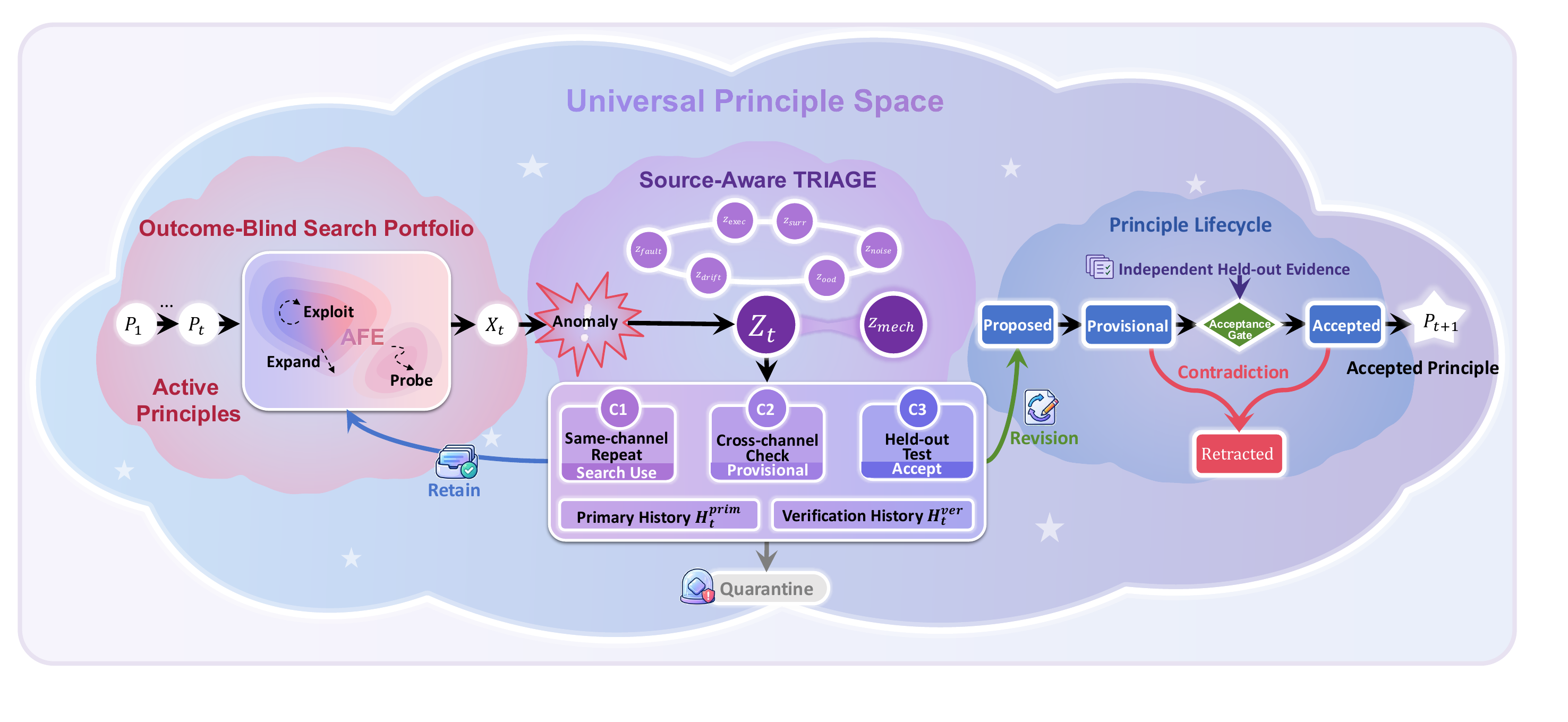}
    \vspace{-0.6cm}
    \caption{\textbf{Overview of \method{}.} \textit{Left:} an outcome-blind AFE portfolio cycles candidate proposals across exploitation, frontier expansion, and mechanism-discriminating probes. \textit{Middle:} \triage{} treats an anomalous residual as source-ambiguous and routes it to quarantine, search history, or verification rather than attributing it to a source at runtime. \textit{Right:} principles advance \textsc{Proposed} $\to$ \textsc{Provisional} $\to$ \textsc{Accepted} only with independent held-out support; contradictory evidence retracts them.}
    \vspace{-0.3cm}
    \label{fig:method-overview}
\end{figure*}

\noindent \textbf{Overview.} \method focuses on maintaining a revisable account of what observations mean. In this context, each experiment has two possible consequences: (a) guiding the next experiment, and (b) changing the principles used to interpret future experiments. Based on our intuition:
\begin{takeaway}
    \centering
    \emph{A surprising observation can steer hypothesizing long before it is reliable enough \\ to revise persistent scientific principles. }
\end{takeaway}
As shown in Figure~\ref{fig:method-overview}, \method{} separates \emph{using an observation} from \emph{believing its explanation}: it keeps exploration responsive to new signals while requiring independent evidence before revising a principle. \method{} builds on the principle-hypothesis-testing cycle of PiEvo~\citep{pu2026principle}.

This separation is implemented as a feedback loop with three components:
(a) an evidence-admission layer checks whether an unexpected outcome generalizes to held-out cases (Section~\ref{sec:verify}),
(b) a search policy proposes experiments from outcomes that have been admitted for exploration (Section~\ref{sec:afe}), and
(c) a principle manager then updates the principle state accordingly (Section~\ref{sec:principles}).
Two histories keep these roles apart: the primary history accumulates observations the search may act on, while the verification history records linked checks that alone can justify---or overturn---a principle revision.

\subsection{Evidence Admission under Source Ambiguity}
\label{sec:revision}

We define the objects on which \triage{} operates, proceeding in three steps: the round-level state, the source-ambiguity model, and the decision objective that any runtime admission rule must respect.

\noindent \textbf{Round state and the anomaly event.}
The agent interacts with an environment over rounds $t=1,\dots,T$, and each round produces exactly one record through three steps: (i) the agent proposes a testable candidate $x_t\in\mathcal{X}$; (ii) an internal predictor---the \emph{Gaussian Process (GP) experts} $\{\mathcal{M}_P\}_{P\in\mathcal{P}_t}$ of \pievo{}~\citep{pu2026principle}, one likelihood model per principle record, fit on the history of admitted observations---issues a prediction $\hat y_t$ with predictive scale $\sigma_t>0$; (iii) an observation channel $c_t\in\mathcal{C}$ (e.g., an evaluator or data source) evaluates $x_t$ and returns an outcome $y_t$ with evaluator's metadata $m_t$ (e.g., configuration). The tuple $(x_t,\hat y_t,y_t,c_t,m_t)$ constitutes the record; how far $y_t$ deviates from what $\mathcal{P}_t$ predicted determines what happens to it next, per Definition~\ref{def:anomaly}.
\begin{defbox}
\begin{definition}[Anomaly score and anomaly event]
    \label{def:anomaly}
    The \textbf{anomaly score} of a record is its standardized residual
    $ s_t \;\coloneqq\; \nicefrac{|y_t-\hat y_t|}{\sigma_t} $.
    Given a threshold $\tau_{\mathrm{anom}}$ fixed before evaluation, the record is ordinary if $s_t\leq\tau_{\mathrm{anom}}$ and an anomaly if $s_t>\tau_{\mathrm{anom}}$.
\end{definition}
\end{defbox}
An ordinary record is admitted directly to the primary history and triggers no further action. An anomaly, by contrast, opens a diagnostic state: it is withheld from search scores and principle updates until the verification procedure of Section~\ref{sec:verify} resolves it.

\noindent \textbf{The anomaly score is blind to its source.}
A large residual is ambiguous in origin. We make this explicit by attaching to each evaluation a source variable $Z_t$ taking values in
\begin{equation}
  \mathcal{Z} \;=\; \underbrace{\{z_{\mathrm{noise}},\, z_{\mathrm{fault}},\, z_{\mathrm{exec}},\, z_{\mathrm{surr}},\, z_{\mathrm{ood}},\, z_{\mathrm{drift}}\}}_{\text{artifact sources } \mathcal{Z}_{\mathrm{art}}} \;\cup\; \underbrace{\{z_{\mathrm{mech}}\}}_{\text{new mechanism}},
  \label{eq:sources}
\end{equation}
denoting measurement noise, device fault, execution failure, mismatch of the external surrogate evaluator, out-of-distribution input, distribution drift, and a genuine new mechanism. Exactly one source generates each evaluation, and only $z_{\mathrm{mech}}$ justifies revision; the remaining \emph{artifact sources} $\mathcal{Z}_{\mathrm{art}}$ should leave the principle state untouched. Because $s_t$ is a deterministic function of $(y_t,\hat y_t,\sigma_t)$ and not of $Z_t$, the anomaly score is uninformative about which source produced it; thresholding $s_t$ alone therefore cannot separate a discovery from an artifact.
The consequences of deciding wrongly are asymmetric, as Example~\ref{ex:asymmetry} illustrates. 

\begin{example}[Two anomalies with identical scores]
\label{ex:asymmetry}
    Suppose two rounds produce the same large score $s_t$. In round $t_1$, a corrupted evaluator channel ($z_{\mathrm{fault}}$) inflates the outcome of a mediocre candidate; revising $\mathcal{P}_t$ on this record redirects every subsequent prediction toward an artifact, corrupting all future rounds. In round $t_2$, a genuinely novel candidate ($z_{\mathrm{mech}}$) violates the current principles; retaining $\mathcal{P}_t$ merely postpones the discovery until stronger evidence arrives. The first error propagates through persistent state; the second only costs time.
\end{example}

To address this asymmetry, we define the admission decision as choosing $d\in\{\retain,\revise\}$ after observing an anomaly, with loss
\begin{equation}
 L(d,z)=\ell_{\mathrm{FR}}\,\mathbb{1}[d=\revise,\,z\in\mathcal{Z}_{\mathrm{art}}] + \ell_{\mathrm{FM}}\,\mathbb{1}[d=\retain,\,z=z_{\mathrm{mech}}],
 \label{eq:loss}
\end{equation}
where $\ell_{\mathrm{FR}}$ and $\ell_{\mathrm{FM}}$ price false revision and false retention, respectively, with $\ell_{\mathrm{FR}}\gg\ell_{\mathrm{FM}}$ per Example~\ref{ex:asymmetry}. Under this loss, the risk-optimal action is to revise only when the posterior $q(z_{\mathrm{mech}}\mid y_t,\hat y_t,m_t)$ exceeds $\ell_{\mathrm{FR}}/(\ell_{\mathrm{FR}}+\ell_{\mathrm{FM}})$---a threshold close to one. 

\begin{takeaway}
    \textbf{Remark.} The \emph{reference model} of Eq.~\eqref{eq:loss} dictates three essential properties: (i) anomaly detection must trigger further evidence collection; (ii) the evidence required for revision must be independent of the channel $c_t$; and (iii) revising persistent state must demand strictly stronger evidence than search use does.
\end{takeaway}

\subsection{Escalating from Repeats to Held-Out Probes}
\label{sec:verify}
\vspace{-2pt}
Because an anomaly may affect candidate search, principle proposal, or principle revision, \triage{} applies three checks (C1$\sim$C3) of increasing independence from the originating experiment channel $c_t$, stopping at the first failure. Each check rules out one class of artifact sources in $\mathcal{Z}_{\mathrm{art}}$ (see Eq.~\eqref{eq:sources}) and unlocks exactly one privilege.

\noindent \textbf{C1: Rules out transient faults; unlocks search use.} Re-evaluating $x_t$ on $c_t$ yields $v_t^{c_t}$. Agreement, $|y_t-v_t^{c_t}|\leq\epsilon_{\mathrm{rep}}$, excludes transient noise and execution failures ($z_{\mathrm{noise}},z_{\mathrm{exec}}$). This cannot authorize revision, since repeating on $c_t$ replicates whatever systematic bias $c_t$ carries.

\noindent \textbf{C2: Rules out channel-specific corruption; unlocks revision proposals.} Re-evaluating $x_t$ on an independent channel $c'\neq c_t$ yields $v_t^{c'}$. Agreement, $|v_t^{c_t}-v_t^{c'}|\leq\epsilon_{\mathrm{ind}}$, excludes corruption specific to $c_t$ ($z_{\mathrm{fault}}$), so the anomaly may propose a principle and become provisional (see Section~\ref{sec:principles}).

\noindent \textbf{C3: Rules out non-generalizing artifacts; unlocks acceptance.} Evaluation by an independent evaluator on held-out inputs excludes effects tied to the original evaluation context ($z_{\mathrm{surr}},z_{\mathrm{ood}},z_{\mathrm{drift}}$), and is the only evidence the acceptance gate counts (see Section~\ref{sec:principles}). A record failing any check is quarantined---excluded from both histories' influence until resolved; a verified replacement may remain search-active but cannot update persistent state by itself.

Reconciliation outcomes update per-channel reliability, so that frequently contradicted channels are progressively excluded from verification duty. Each channel $c$ carries a reliability estimate $r_c\sim\operatorname{Beta}(\alpha_c,\beta_c)$ with trust score $p_c^{\mathrm{trust}}=\alpha_c/(\alpha_c+\beta_c)$; agreement increments $\alpha_c$ and disagreement increments $\beta_c$ for every participating channel (see Algorithm~\ref{alg:triage}).
\vspace{-3pt}

\begin{takeaway}
    \noindent \textbf{Remark.} Trust selects which channels may verify, but does not lower the standard of evidence. All thresholds are fixed before evaluation and are independent of $p_c^{\mathrm{trust}}$, so no amount of accumulated trust lets a channel certify its own anomalies.
\end{takeaway}
\vspace{-4pt}

\subsection{Proposing Candidates without Seeing the Gate}
\label{sec:afe}
\vspace{-2pt}
\begin{wrapfigure}[18]{r}{0.37\textwidth}
    \vspace{-14pt}
    \centering
    \includegraphics[width=\linewidth]{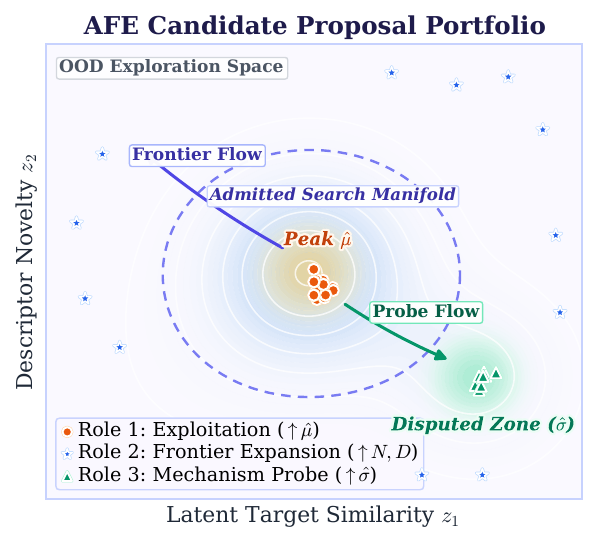}
    \vspace{-15pt}
    \caption{\textbf{AFE portfolio schematic.} Three complementary roles on a shared design manifold: exploitation ($\uparrow\hat\mu$), frontier expansion ($\uparrow N, D$), and mechanism probes ($\uparrow\hat\sigma$).}
    \label{fig:afe-manifold}
    \vspace{-12pt}
\end{wrapfigure}
We use Active Flow Expansion (AFE)~\citep{de2026active} as a search component in \method{}. It proposes candidates in \emph{three roles}: best-region \textbf{exploitation}, frontier \textbf{expansion}, and mechanism-discriminating \textbf{probes}.

\noindent \textbf{Scoring candidates.}
For each role $r$, AFE scores a candidate $x\in\mathcal{X}$ by four terms computed from the GP experts of Section~\ref{sec:revision}, refit on the admitted primary history:
\[
A_r(x) \;=\; w_r^{\mu}\,\hat{\mu}(x) \;+\; w_r^{\sigma}\,\hat{\sigma}(x) \;+\; w_r^{n}\,N(x) \;+\; w_r^{d}\,D(x),
\]
where $\hat\mu(x)$ and $\hat\sigma(x)$ are the GP experts' predicted outcome and predictive uncertainty, $N(x)$ is the distance of $x$ from already evaluated inputs, and $D(x)$ is a descriptor-novelty measure. The role weights $\{w_r^{\mu},w_r^{\sigma},w_r^{n},w_r^{d}\}$ are fixed a priori and differ across roles: exploitation weights $\hat\mu$, expansion weights $N$ and $D$, and probes weight $\hat\sigma$ in regions where competing principles disagree (Figure~\ref{fig:afe-manifold}).

\noindent \textbf{AFE reads admitted evidence; \triage{} alone writes it.}
Two append-only histories keep search evidence and revision evidence apart: the \emph{primary history} $H_t^{\mathrm{prim}}$ stores the records the search policy may act on, and the \emph{verification history} $H_t^{\mathrm{ver}}$ stores the C1--C3 checks (see Section~\ref{sec:verify}), each linked to the primary record it concerns. ``Outcome-blind'' is then a one-way boundary over these histories: AFE reads only $H_t^{\mathrm{prim}}$ and candidate descriptors, outputs $x_t$, and is forbidden from verification channels, quarantine flags, and gate state; \triage{} alone writes back, entering the admitted, substituted, or quarantined outcome into $H_{t+1}^{\mathrm{prim}}$ for the next refit of the GP experts---\pievo's per-principle likelihood models $p(y\mid h,P)$~\citep{pu2026principle}, which we use unmodified. Unable to observe gate state, the search policy can never smuggle unverified evidence into persistent state; the only write path to what the agent believes is the acceptance gate of Section~\ref{sec:principles}.

\subsection{A Gated Principle Lifecycle with Reversible Revision}
\label{sec:principles}

The final component determines how verified evidence changes persistent state. Each principle $P$ is a record $(\mathrm{parent}(P),E(P),\mathrm{state}(P))$---a pointer to the principle it refines, its linked evidence set, and a lifecycle state that advances through
$
\textsc{Proposed}\;\longrightarrow\;\textsc{Provisional}\;\longrightarrow\;\textsc{Accepted},
$
with any state sent to \textsc{Retracted} upon contradiction. We specify the two forward transitions and the retraction rule in turn.

\noindent \textbf{From proposal to acceptance: only admissible evidence counts.}
An anomalous record passing C1--C2 (see Section~\ref{sec:verify}) creates or updates a \textsc{Proposed} principle, and C2's cross-channel agreement promotes it to \textsc{Provisional}. Acceptance, however, is gated by the following class of evidence:

\begin{defbox}
\begin{definition}[Admissible evidence for revision]
\label{def:admissible}
    A record $e$ is \emph{admissible} for principle $P$ if (i) it is produced by an evaluator process sharing no memory, cache, or random seed with the generator (\emph{independence}), and (ii) it is measured on $X_{\mathrm{heldout}}$ disjoint from the generator-visible $X_{\mathrm{train}}$ (\emph{held-out}). An admissible record \emph{supports} $P$ if its held-out gain $\Delta_{\mathrm{heldout}}(e;P)\geq\tau_{\mathrm{gen}}$ and \emph{contradicts} $P$ if $\Delta_{\mathrm{heldout}}(e;P)\leq-\tau_{\mathrm{gen}}$; records in between are uninformative.
\end{definition}
\end{defbox}

Writing $n_P^{+}$ and $n_P^{-}$ for the numbers of supporting and contradicting admissible records in $E(P)$,
$
 \textsc{Provisional}\;\longrightarrow\;\textsc{Accepted}
 \Longleftrightarrow 
 n_P^{-}=0\ \ \wedge\ \ n_P^{+}\geq k .
$
The two conditions of Definition~\ref{def:admissible} each remove one failure mode: independence prevents corroboration from being manufactured by the proposing channel, and the held-out requirement makes acceptance test generalization rather than fit. 

\noindent \textbf{Retraction makes revision reversible.}
A single contradicting admissible record ($n_P^{-}\geq 1$) moves any active principle---\textsc{Proposed}, \textsc{Provisional}, or \textsc{Accepted}---to \textsc{Retracted}, and the retraction propagates through $\mathrm{parent}(\cdot)$, so descendants of a contradicted principle cannot remain trusted. Retraction is standing rather than proposal-time only: every newly admitted record is scored against each active principle it bears on, so evidence arriving after acceptance can still overturn an \textsc{Accepted} principle (Algorithm~\ref{alg:triage}).

In summary, the full loop of \method, i.e., AFE proposal, evaluation, staged verification (C1--C3), trust update, and lifecycle transition, is summarized in Algorithm~\ref{alg:triage}, with thresholds and sensitivity analyses in Appendix~\ref{app:hyperparams}.

\section{Experiments}
\label{sec:experiments}

We evaluate \method{} along four axes: (1) matched downstream performance across four scientific discovery tasks under three backbone LLMs; (2) search efficiency in optimization speed, candidate coverage, and computational cost; (3) controlled stress tests of the admission gate under evaluator corruption, hidden anomaly sources, mechanism reversal, proxy exploitation, and cross-harness deployment; and (4) ablations isolating search expansion from evidence admission.

\subsection{Experiment Setup}
\label{sec:matched}

\noindent \textbf{Tasks and Benchmarks.} Following~\citet{pu2026principle}, we evaluate \method{} on four optimization tasks: Molecular Bio-activity (MBO), Nanohelix (NHO), Superconductor (SPO), and Transition Metal Complex (TMC). Each task is scored by its domain evaluator and maximizes a task-specific objective under a shared budget of $24$ evaluations.

\noindent \textbf{Baselines.} We evaluate three categories of scientific agents : (a) \textbf{generic and classic}: Vanilla MAS and ReAct~\citep{yao2022react}, which represent tool-use loops without persistent scientific state; (b) \textbf{autonomous research}: AI Scientist v1~\citep{lu2024ai}, AI Scientist v2~\citep{yamada2025ai}, and AI Researcher~\citep{tang2026ai}, which automate multi-stage pipelines over transient artifacts; and (c) \textbf{principle-evolving}: PiFlow~\citep{pu2025piflow} and \pievo{}~\citep{pu2026principle}, the state of the art that revises a principle set by anomalous---precisely the response our method re-governs.

\noindent \textbf{Implementation Details.} We evaluate three backbones: Qwen3-235B~\citep{qwen3technicalreport}, Gemini-2.5-Flash-Lite~\citep{googlegemini25flashlite}, GPT-5.6-Luna~\citep{openai_gpt56} without extended chain-of-thought, over three fixed seeds, reported as mean $\pm$ std. Gate parameters in \method are fixed a priori across all tasks and backbones. Full specification and sensitivity sweeps in Appendix~\ref{app:hyperparams}.


\subsection{Evaluation Metrics}
\label{sec:metrics}
With $y_{1:T}$ the primary outcomes of a run, we employ metrics that capture abilities to (a) reach high-quality solutions, (b) reach them early, and (c) cover the candidate space broadly:

\noindent \textbf{a. Solution Quality (SQ, \%, $\uparrow$).} The best outcome found, normalized by a published per-task reference $\mu_{\mathrm{ref}}$ from \pievo{}~\citep{pu2026principle}: $\mathrm{SQ}=100\cdot\nicefrac{\max_t y_t}{\mu_{\mathrm{ref}}}$, with $\mu_{\mathrm{ref}}=6.5$ (MBO), $2.0$ (NHO), $298.5$ (SPO), and $493.8$ (TMC). These references are not attainable upper bounds, so SQ passes $100$ when a run beats the reference; we apply no clipping.

\noindent \textbf{b. Area Under the Optimization Curve (AUOC, \%, $\uparrow$).} The budget-averaged best-so-far quality, $\mathrm{AUOC}=100\cdot\frac{1}{T}\sum_{t}\nicefrac{\max_{t'\leq t} y_{t'}}{\mu_{\mathrm{ref}}}$, crediting how early quality is reached.

\noindent \textbf{c. Average Pairwise Distance (APD, $\uparrow$).} The mean pairwise cosine distance between proposed candidates. A higher APD indicates broader candidate coverage.

\begin{table*}[t!]
\caption{\textbf{Solution Quality (SQ, \%) $\uparrow$ across four discovery tasks and three backbone LLMs. } Entries report mean $\pm$ std over three seeds; \textbf{Bold} indicates the best result within each backbone.}
\vspace{-0.2cm}
\label{tab:main-sq}
\centering
\small
\resizebox{\linewidth}{!}{%
\begin{tabular}{l | cccc | c}
\toprule
\textbf{Model / Method} & \textbf{MBO} & \textbf{NHO} & \textbf{SPO} & \textbf{TMC} & \textbf{Average} \\
\midrule
\multicolumn{6}{l}{\textbf{Qwen3-235B (No Thinking)}} \\
\midrule
Vanilla MAS & 83.35 \textcolor{gray}{\scriptsize $\pm$ 29.95} & 52.83 \textcolor{gray}{\scriptsize $\pm$ 20.36} & 37.29 \textcolor{gray}{\scriptsize $\pm$ 0.75} & 56.34 \textcolor{gray}{\scriptsize $\pm$ 3.69} & 57.45  \\
ReAct & 75.48 \textcolor{gray}{\scriptsize $\pm$ 4.94} \textcolor{red}{\scriptsize $\downarrow$ 0.9x} & 61.95 \textcolor{gray}{\scriptsize $\pm$ 17.08} \textcolor{purple}{\scriptsize $\uparrow$ 1.2x} & 37.49 \textcolor{gray}{\scriptsize $\pm$ 0.46} & 72.80 \textcolor{gray}{\scriptsize $\pm$ 4.48} \textcolor{purple}{\scriptsize $\uparrow$ 1.3x} & 61.93  \textcolor{purple}{\scriptsize $\uparrow$ 1.1x} \\
AI Scientist v1 & 73.37 \textcolor{gray}{\scriptsize $\pm$ 39.82} \textcolor{red}{\scriptsize $\downarrow$ 0.9x} & 41.50 \textcolor{gray}{\scriptsize $\pm$ 3.12} \textcolor{red}{\scriptsize $\downarrow$ 0.8x} & 37.04 \textcolor{gray}{\scriptsize $\pm$ 0.61} & 63.96 \textcolor{gray}{\scriptsize $\pm$ 8.93} \textcolor{purple}{\scriptsize $\uparrow$ 1.1x} & 53.97  \textcolor{red}{\scriptsize $\downarrow$ 0.9x} \\
AI Scientist v2 & 84.40 \textcolor{gray}{\scriptsize $\pm$ 20.07} & 52.34 \textcolor{gray}{\scriptsize $\pm$ 15.56} & 41.40 \textcolor{gray}{\scriptsize $\pm$ 1.15} \textcolor{purple}{\scriptsize $\uparrow$ 1.1x} & 63.52 \textcolor{gray}{\scriptsize $\pm$ 2.86} \textcolor{purple}{\scriptsize $\uparrow$ 1.1x} & 60.42  \textcolor{purple}{\scriptsize $\uparrow$ 1.1x} \\
AI Researcher & 105.70 \textcolor{gray}{\scriptsize $\pm$ 21.92} \textcolor{purple}{\scriptsize $\uparrow$ 1.3x} & 37.21 \textcolor{gray}{\scriptsize $\pm$ 5.68} \textcolor{red}{\scriptsize $\downarrow$ 0.7x} & 37.39 \textcolor{gray}{\scriptsize $\pm$ 0.85} & 78.80 \textcolor{gray}{\scriptsize $\pm$ 3.76} \textcolor{purple}{\scriptsize $\uparrow$ 1.4x} & 64.78  \textcolor{purple}{\scriptsize $\uparrow$ 1.1x} \\
PiFlow & 81.89 \textcolor{gray}{\scriptsize $\pm$ 30.49} & 56.15 \textcolor{gray}{\scriptsize $\pm$ 15.23} \textcolor{purple}{\scriptsize $\uparrow$ 1.1x} & 37.21 \textcolor{gray}{\scriptsize $\pm$ 0.52} & 54.58 \textcolor{gray}{\scriptsize $\pm$ 2.70} & 57.46  \\
\pievo{} & 97.84 \textcolor{gray}{\scriptsize $\pm$ 27.49} \textcolor{purple}{\scriptsize $\uparrow$ 1.2x} & 65.53 \textcolor{gray}{\scriptsize $\pm$ 21.90} \textcolor{purple}{\scriptsize $\uparrow$ 1.2x} & 37.71 \textcolor{gray}{\scriptsize $\pm$ 0.53} & 80.30 \textcolor{gray}{\scriptsize $\pm$ 4.57} \textcolor{purple}{\scriptsize $\uparrow$ 1.4x} & 70.35  \textcolor{purple}{\scriptsize $\uparrow$ 1.2x} \\
\textbf{Triage-PiEvo (ours)} & \textbf{132.00} \textcolor{gray}{\scriptsize $\pm$ 9.84} \textcolor{purple}{\scriptsize $\uparrow$ 1.6x} & \textbf{72.89} \textcolor{gray}{\scriptsize $\pm$ 17.80} \textcolor{purple}{\scriptsize $\uparrow$ 1.4x} & \textbf{42.00} \textcolor{gray}{\scriptsize $\pm$ 1.20} \textcolor{purple}{\scriptsize $\uparrow$ 1.1x} & \textbf{81.50} \textcolor{gray}{\scriptsize $\pm$ 4.02} \textcolor{purple}{\scriptsize $\uparrow$ 1.4x} & \textbf{82.10}  \textcolor{purple}{\scriptsize $\uparrow$ 1.4x} \\
\midrule
\multicolumn{6}{l}{\textbf{Gemini-2.5-Flash-Lite (No Thinking)}} \\
\midrule
Vanilla MAS & 62.06 \textcolor{gray}{\scriptsize $\pm$ 14.09} & 71.98 \textcolor{gray}{\scriptsize $\pm$ 1.01} & 36.95 \textcolor{gray}{\scriptsize $\pm$ 0.46} & 66.64 \textcolor{gray}{\scriptsize $\pm$ 17.42} & 59.41  \\
ReAct & 58.83 \textcolor{gray}{\scriptsize $\pm$ 6.44} \textcolor{red}{\scriptsize $\downarrow$ 0.9x} & 69.90 \textcolor{gray}{\scriptsize $\pm$ 1.50} & 32.06 \textcolor{gray}{\scriptsize $\pm$ 5.75} \textcolor{red}{\scriptsize $\downarrow$ 0.9x} & 56.31 \textcolor{gray}{\scriptsize $\pm$ 2.37} \textcolor{red}{\scriptsize $\downarrow$ 0.8x} & 54.27  \textcolor{red}{\scriptsize $\downarrow$ 0.9x} \\
AI Scientist v1 & 75.60 \textcolor{gray}{\scriptsize $\pm$ 22.03} \textcolor{purple}{\scriptsize $\uparrow$ 1.2x} & 73.58 \textcolor{gray}{\scriptsize $\pm$ 2.23} & 37.86 \textcolor{gray}{\scriptsize $\pm$ 2.55} & 55.70 \textcolor{gray}{\scriptsize $\pm$ 2.62} \textcolor{red}{\scriptsize $\downarrow$ 0.8x} & 60.68  \\
AI Scientist v2 & 99.30 \textcolor{gray}{\scriptsize $\pm$ 22.73} \textcolor{purple}{\scriptsize $\uparrow$ 1.6x} & 74.87 \textcolor{gray}{\scriptsize $\pm$ 2.00} & 30.74 \textcolor{gray}{\scriptsize $\pm$ 3.96} \textcolor{red}{\scriptsize $\downarrow$ 0.8x} & 54.80 \textcolor{gray}{\scriptsize $\pm$ 4.09} \textcolor{red}{\scriptsize $\downarrow$ 0.8x} & 64.93  \textcolor{purple}{\scriptsize $\uparrow$ 1.1x} \\
AI Researcher & 66.47 \textcolor{gray}{\scriptsize $\pm$ 16.70} \textcolor{purple}{\scriptsize $\uparrow$ 1.1x} & 65.39 \textcolor{gray}{\scriptsize $\pm$ 9.16} \textcolor{red}{\scriptsize $\downarrow$ 0.9x} & 31.62 \textcolor{gray}{\scriptsize $\pm$ 3.82} \textcolor{red}{\scriptsize $\downarrow$ 0.9x} & 63.46 \textcolor{gray}{\scriptsize $\pm$ 13.85} & 56.74  \\
PiFlow & 96.59 \textcolor{gray}{\scriptsize $\pm$ 11.71} \textcolor{purple}{\scriptsize $\uparrow$ 1.6x} & 74.24 \textcolor{gray}{\scriptsize $\pm$ 4.06} & 34.91 \textcolor{gray}{\scriptsize $\pm$ 3.66} \textcolor{red}{\scriptsize $\downarrow$ 0.9x} & 68.22 \textcolor{gray}{\scriptsize $\pm$ 8.71} & 68.49  \textcolor{purple}{\scriptsize $\uparrow$ 1.2x} \\
\pievo{} & 72.59 \textcolor{gray}{\scriptsize $\pm$ 6.70} \textcolor{purple}{\scriptsize $\uparrow$ 1.2x} & 81.00 \textcolor{gray}{\scriptsize $\pm$ 3.06} \textcolor{purple}{\scriptsize $\uparrow$ 1.1x} & 38.02 \textcolor{gray}{\scriptsize $\pm$ 0.50} & 78.50 \textcolor{gray}{\scriptsize $\pm$ 7.60} \textcolor{purple}{\scriptsize $\uparrow$ 1.2x} & 67.53  \textcolor{purple}{\scriptsize $\uparrow$ 1.1x} \\
\textbf{Triage-PiEvo (ours)} & \textbf{115.00} \textcolor{gray}{\scriptsize $\pm$ 10.00} \textcolor{purple}{\scriptsize $\uparrow$ 1.9x} & \textbf{83.00} \textcolor{gray}{\scriptsize $\pm$ 5.00} \textcolor{purple}{\scriptsize $\uparrow$ 1.2x} & \textbf{40.50} \textcolor{gray}{\scriptsize $\pm$ 4.00} \textcolor{purple}{\scriptsize $\uparrow$ 1.1x} & \textbf{80.00} \textcolor{gray}{\scriptsize $\pm$ 4.00} \textcolor{purple}{\scriptsize $\uparrow$ 1.2x} & \textbf{79.63}  \textcolor{purple}{\scriptsize $\uparrow$ 1.3x} \\
\midrule
\multicolumn{6}{l}{\textbf{GPT-5.6-Luna (No Thinking)}} \\
\midrule
Vanilla MAS & 91.84 \textcolor{gray}{\scriptsize $\pm$ 18.08} & 66.26 \textcolor{gray}{\scriptsize $\pm$ 7.14} & 36.42 \textcolor{gray}{\scriptsize $\pm$ 0.50} & 68.07 \textcolor{gray}{\scriptsize $\pm$ 3.91} & 65.65  \\
ReAct & 132.00 \textcolor{gray}{\scriptsize $\pm$ 22.98} \textcolor{purple}{\scriptsize $\uparrow$ 1.4x} & 69.98 \textcolor{gray}{\scriptsize $\pm$ 0.14} \textcolor{purple}{\scriptsize $\uparrow$ 1.1x} & 36.22 \textcolor{gray}{\scriptsize $\pm$ 0.34} & 65.37 \textcolor{gray}{\scriptsize $\pm$ 5.39} & 75.89  \textcolor{purple}{\scriptsize $\uparrow$ 1.2x} \\
AI Scientist v1 & 91.16 \textcolor{gray}{\scriptsize $\pm$ 19.35} & 53.02 \textcolor{gray}{\scriptsize $\pm$ 12.32} \textcolor{red}{\scriptsize $\downarrow$ 0.8x} & 35.52 \textcolor{gray}{\scriptsize $\pm$ 4.90} & 62.44 \textcolor{gray}{\scriptsize $\pm$ 6.24} \textcolor{red}{\scriptsize $\downarrow$ 0.9x} & 60.54  \textcolor{red}{\scriptsize $\downarrow$ 0.9x} \\
AI Scientist v2 & 89.63 \textcolor{gray}{\scriptsize $\pm$ 31.45} & 65.13 \textcolor{gray}{\scriptsize $\pm$ 15.46} & 36.77 \textcolor{gray}{\scriptsize $\pm$ 5.05} & 58.29 \textcolor{gray}{\scriptsize $\pm$ 0.95} \textcolor{red}{\scriptsize $\downarrow$ 0.9x} & 62.46  \\
AI Researcher & 77.86 \textcolor{gray}{\scriptsize $\pm$ 15.38} \textcolor{red}{\scriptsize $\downarrow$ 0.8x} & 73.92 \textcolor{gray}{\scriptsize $\pm$ 9.01} \textcolor{purple}{\scriptsize $\uparrow$ 1.1x} & 36.28 \textcolor{gray}{\scriptsize $\pm$ 0.24} & 72.93 \textcolor{gray}{\scriptsize $\pm$ 2.06} \textcolor{purple}{\scriptsize $\uparrow$ 1.1x} & 65.25  \\
PiFlow & 128.77 \textcolor{gray}{\scriptsize $\pm$ 18.74} \textcolor{purple}{\scriptsize $\uparrow$ 1.4x} & 46.62 \textcolor{gray}{\scriptsize $\pm$ 17.77} \textcolor{red}{\scriptsize $\downarrow$ 0.7x} & 31.49 \textcolor{gray}{\scriptsize $\pm$ 5.94} \textcolor{red}{\scriptsize $\downarrow$ 0.9x} & 59.59 \textcolor{gray}{\scriptsize $\pm$ 5.23} \textcolor{red}{\scriptsize $\downarrow$ 0.9x} & 66.62  \\
\pievo{} & 69.75 \textcolor{gray}{\scriptsize $\pm$ 6.20} \textcolor{red}{\scriptsize $\downarrow$ 0.8x} & 67.82 \textcolor{gray}{\scriptsize $\pm$ 10.87} & 37.41 \textcolor{gray}{\scriptsize $\pm$ 0.53} & 76.50 \textcolor{gray}{\scriptsize $\pm$ 1.78} \textcolor{purple}{\scriptsize $\uparrow$ 1.1x} & 62.87  \\
\textbf{Triage-PiEvo (ours)} & \textbf{136.00} \textcolor{gray}{\scriptsize $\pm$ 9.84} \textcolor{purple}{\scriptsize $\uparrow$ 1.5x} & \textbf{78.18} \textcolor{gray}{\scriptsize $\pm$ 5.44} \textcolor{purple}{\scriptsize $\uparrow$ 1.2x} & \textbf{40.80} \textcolor{gray}{\scriptsize $\pm$ 5.70} \textcolor{purple}{\scriptsize $\uparrow$ 1.1x} & \textbf{78.50} \textcolor{gray}{\scriptsize $\pm$ 3.75} \textcolor{purple}{\scriptsize $\uparrow$ 1.2x} & \textbf{83.37}  \textcolor{purple}{\scriptsize $\uparrow$ 1.3x} \\
\bottomrule
\end{tabular}%
}
\end{table*}

\begin{figure*}[t!]
    \centering
    \vspace{-0.35cm}
    \includegraphics[width=\textwidth]{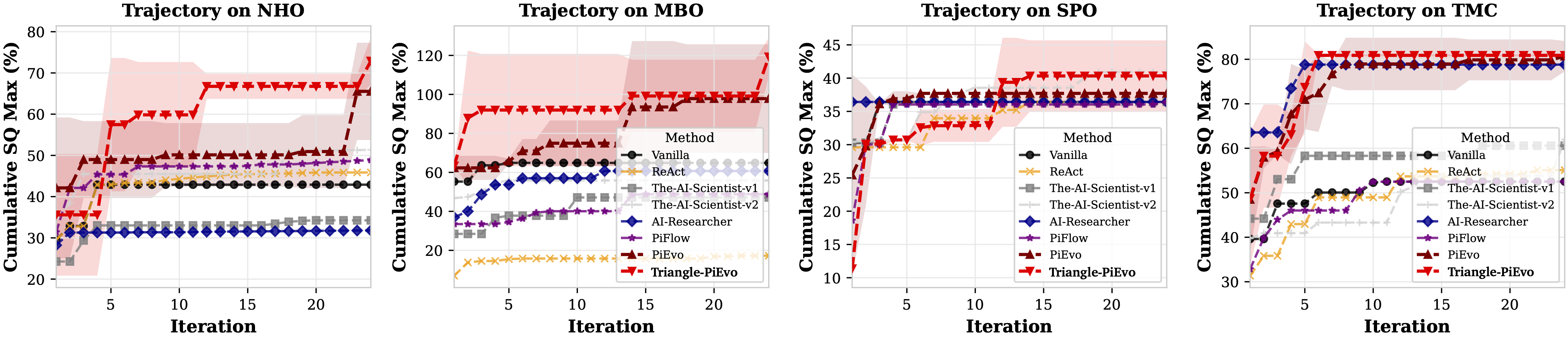}
    \vspace{-0.65cm}
    \caption{\textbf{Best-so-far trajectories across four benchmarks with Qwen3-235B backbone.} Across all scientific domains, \method{} accumulates high-quality candidates and sustains the highest solution quality.}
    \label{fig:qwen-convergence}
    \vspace{-0.45cm}
\end{figure*}

\subsection{Performance and Efficiency Analysis}
\label{sec:performance}

Table~\ref{tab:main-sq} compares \method{} against seven baselines on solution quality (SQ) across four tasks and three backbones, and Table~\ref{tab:main-apd-auoc} reports the corresponding coverage (APD) and accumulation speed (AUOC). Our analysis yields three primary observations:

\noindent \textbf{Obs.\ding{202}: \method attains the highest average SQ under every backbone.} 
\method{} attains the highest SQ in every task–backbone and the highest average under every backbone, improving on the strongest baseline of each backbone by $11.75\%$, $11.14\%$, and $7.48\%$ (Table~\ref{tab:main-sq}), indicating that evidence admission repairs a failure mode shared across three backbones rather than one tuned to a single model.

\noindent \textbf{Obs.\ding{203}: \method{} leads on average coverage and on accumulation speed.}
\method achieves a $1.16\times$--$1.22\times$ faster accumulation (Table~\ref{tab:main-apd-auoc}), while baselines trade the two facets off (e.g., AI Researcher~\citep{tang2026ai} reaches $89.21$ AUOC on MBO but at the lowest average coverage); on NHO, ReAct and AI Scientist v2 attain higher coverage but lower AUOC, so the lead holds on average rather than in every task--metric cell. This confirms that \method filters evidence for revision without constricting the search.


\noindent \textbf{Obs.\ding{204}: \method opens its margin early and holds it at a bounded cost.} 
On Qwen3-235B, \method buys the strongest baseline's full-budget result using one-third of the budget (Figure~\ref{fig:qwen-convergence}; Appendix~\ref{app:apd-auoc}, Figure~\ref{fig:budget-pareto}). The gate adds only $+13.6\%$ wall-clock time over \pievo{} ($134.5$s vs.\ $118.4$s per run), while the multi-agent and tree-search frameworks it outperforms cost $1.28\times$--$1.85\times$ more (AI Scientist v2: $172.4$s; AI Researcher: $188.2$s; PiFlow: $248.6$s), placing \method{} on the efficiency frontier of quality versus cost (Appendix~\ref{app:runtime}, Figure~\ref{fig:runtime}). 


\begin{table*}[t!]
\caption{\textbf{Average Pairwise Distance (APD, $\uparrow$) and Area Under the Optimisation Curve (AUOC, $\uparrow$) across four tasks and three backbone LLMs.} Higher APD indicates broader candidate coverage, while higher AUOC indicates faster accumulation of high-quality candidates.}
\label{tab:main-apd-auoc}
\centering
\small
\setlength{\tabcolsep}{2pt}
\resizebox{\linewidth}{!}{%
\begin{tabular}{l | cc | cc | cc | cc | cc}
\toprule
\textbf{Model / Method} & \multicolumn{2}{c}{\textbf{MBO}} & \multicolumn{2}{c}{\textbf{NHO}} & \multicolumn{2}{c}{\textbf{SPO}} & \multicolumn{2}{c}{\textbf{TMC}} & \multicolumn{2}{c}{\textbf{Average}} \\
 & \textbf{APD} $\uparrow$ & \textbf{AUOC} $\uparrow$ & \textbf{APD} $\uparrow$ & \textbf{AUOC} $\uparrow$ & \textbf{APD} $\uparrow$ & \textbf{AUOC} $\uparrow$ & \textbf{APD} $\uparrow$ & \textbf{AUOC} $\uparrow$ & \textbf{Avg APD} $\uparrow$ & \textbf{Avg AUOC} $\uparrow$ \\
\midrule
\multicolumn{11}{l}{\textbf{Qwen3-235B (No Thinking)}} \\
\midrule
Vanilla MAS & 53.58\textcolor{gray}{\tiny$\pm$15.25} & 70.09\textcolor{gray}{\tiny$\pm$17.75} & 32.23\textcolor{gray}{\tiny$\pm$3.72} & 49.33\textcolor{gray}{\tiny$\pm$17.82} & 52.03\textcolor{gray}{\tiny$\pm$1.05} & 36.34\textcolor{gray}{\tiny$\pm$0.38} & 51.89\textcolor{gray}{\tiny$\pm$3.85} & 52.67\textcolor{gray}{\tiny$\pm$3.49} & 47.43 & 52.11 \\
ReAct & 59.43\textcolor{gray}{\tiny$\pm$8.60} & 70.58\textcolor{gray}{\tiny$\pm$7.95} & \textbf{33.31}\textcolor{gray}{\tiny$\pm$1.86} & 47.46\textcolor{gray}{\tiny$\pm$7.04} & 56.13\textcolor{gray}{\tiny$\pm$4.49} & 35.74\textcolor{gray}{\tiny$\pm$0.76} & 72.93\textcolor{gray}{\tiny$\pm$0.33} & 66.95\textcolor{gray}{\tiny$\pm$4.67} & 55.45 \textcolor{purple}{\small $\uparrow$ 1.2x} & 55.18 \textcolor{purple}{\small $\uparrow$ 1.1x} \\
AI Scientist v1 & 52.93\textcolor{gray}{\tiny$\pm$15.22} & 56.16\textcolor{gray}{\tiny$\pm$26.05} & 32.98\textcolor{gray}{\tiny$\pm$2.32} & 38.84\textcolor{gray}{\tiny$\pm$2.06} & 42.55\textcolor{gray}{\tiny$\pm$5.61} & 35.62\textcolor{gray}{\tiny$\pm$1.46} & 64.43\textcolor{gray}{\tiny$\pm$7.02} & 57.50\textcolor{gray}{\tiny$\pm$6.15} & 48.22 & 47.03 \textcolor{red}{\small $\downarrow$ 0.9x} \\
AI Scientist v2 & 60.26\textcolor{gray}{\tiny$\pm$10.57} & 76.03\textcolor{gray}{\tiny$\pm$7.92} & 28.39\textcolor{gray}{\tiny$\pm$3.49} & 39.51\textcolor{gray}{\tiny$\pm$3.29} & 48.18\textcolor{gray}{\tiny$\pm$1.41} & 37.03\textcolor{gray}{\tiny$\pm$0.22} & 55.66\textcolor{gray}{\tiny$\pm$2.28} & 54.75\textcolor{gray}{\tiny$\pm$6.66} & 48.12 & 51.83 \\
AI Researcher & 59.17\textcolor{gray}{\tiny$\pm$3.77} & 89.21\textcolor{gray}{\tiny$\pm$19.08} & 25.51\textcolor{gray}{\tiny$\pm$3.45} & 34.51\textcolor{gray}{\tiny$\pm$3.80} & 47.62\textcolor{gray}{\tiny$\pm$3.55} & 36.82\textcolor{gray}{\tiny$\pm$0.41} & 52.46\textcolor{gray}{\tiny$\pm$6.03} & 74.19\textcolor{gray}{\tiny$\pm$0.70} & 46.19 & 58.68 \textcolor{purple}{\small $\uparrow$ 1.1x} \\
PiFlow & 64.99\textcolor{gray}{\tiny$\pm$2.39} & 64.04\textcolor{gray}{\tiny$\pm$14.87} & 28.07\textcolor{gray}{\tiny$\pm$1.85} & 52.93\textcolor{gray}{\tiny$\pm$15.82} & 51.24\textcolor{gray}{\tiny$\pm$4.14} & 35.90\textcolor{gray}{\tiny$\pm$1.33} & 57.76\textcolor{gray}{\tiny$\pm$3.35} & 50.26\textcolor{gray}{\tiny$\pm$1.85} & 50.52 \textcolor{purple}{\small $\uparrow$ 1.1x} & 50.78 \\
\pievo{} & 63.41\textcolor{gray}{\tiny$\pm$10.78} & 84.14\textcolor{gray}{\tiny$\pm$16.22} & 23.80\textcolor{gray}{\tiny$\pm$5.23} & 52.50\textcolor{gray}{\tiny$\pm$13.36} & 54.41\textcolor{gray}{\tiny$\pm$1.42} & 36.83\textcolor{gray}{\tiny$\pm$0.86} & 71.76\textcolor{gray}{\tiny$\pm$2.58} & 74.93\textcolor{gray}{\tiny$\pm$5.05} & 53.35 \textcolor{purple}{\small $\uparrow$ 1.1x} & 62.10 \textcolor{purple}{\small $\uparrow$ 1.2x} \\
\textbf{Triage-PiEvo (ours)} & \textbf{73.71}\textcolor{gray}{\tiny$\pm$2.86} & \textbf{110.00}\textcolor{gray}{\tiny$\pm$18.41} & 23.67\textcolor{gray}{\tiny$\pm$0.79} & \textbf{61.21}\textcolor{gray}{\tiny$\pm$11.52} & \textbf{57.34}\textcolor{gray}{\tiny$\pm$0.26} & \textbf{38.50}\textcolor{gray}{\tiny$\pm$1.80} & \textbf{74.20}\textcolor{gray}{\tiny$\pm$1.41} & \textbf{77.20}\textcolor{gray}{\tiny$\pm$2.95} & \textbf{57.23} \textcolor{purple}{\small $\uparrow$ 1.2x} & \textbf{71.73} \textcolor{purple}{\small $\uparrow$ 1.4x} \\
\midrule
\multicolumn{11}{l}{\textbf{Gemini-2.5-Flash-Lite (No Thinking)}} \\
\midrule
Vanilla MAS & 65.06\textcolor{gray}{\tiny$\pm$5.46} & 46.01\textcolor{gray}{\tiny$\pm$15.24} & 33.55\textcolor{gray}{\tiny$\pm$0.90} & 61.71\textcolor{gray}{\tiny$\pm$2.41} & 53.65\textcolor{gray}{\tiny$\pm$1.10} & 35.45\textcolor{gray}{\tiny$\pm$0.05} & 57.23\textcolor{gray}{\tiny$\pm$5.29} & 62.88\textcolor{gray}{\tiny$\pm$15.97} & 52.37 & 51.51 \\
ReAct & 62.08\textcolor{gray}{\tiny$\pm$1.15} & 44.54\textcolor{gray}{\tiny$\pm$2.54} & 34.46\textcolor{gray}{\tiny$\pm$0.57} & 58.08\textcolor{gray}{\tiny$\pm$5.33} & 42.17\textcolor{gray}{\tiny$\pm$12.13} & 25.95\textcolor{gray}{\tiny$\pm$8.19} & 52.75\textcolor{gray}{\tiny$\pm$5.72} & 52.03\textcolor{gray}{\tiny$\pm$0.12} & 47.87 \textcolor{red}{\small $\downarrow$ 0.9x} & 45.15 \textcolor{red}{\small $\downarrow$ 0.9x} \\
AI Scientist v1 & 52.02\textcolor{gray}{\tiny$\pm$15.45} & 58.73\textcolor{gray}{\tiny$\pm$5.53} & 33.13\textcolor{gray}{\tiny$\pm$0.89} & 64.99\textcolor{gray}{\tiny$\pm$5.57} & 47.13\textcolor{gray}{\tiny$\pm$4.98} & 33.96\textcolor{gray}{\tiny$\pm$1.93} & 56.53\textcolor{gray}{\tiny$\pm$1.71} & 51.60\textcolor{gray}{\tiny$\pm$0.09} & 47.20 \textcolor{red}{\small $\downarrow$ 0.9x} & 52.32 \\
AI Scientist v2 & 46.27\textcolor{gray}{\tiny$\pm$12.65} & 59.35\textcolor{gray}{\tiny$\pm$11.81} & \textbf{35.09}\textcolor{gray}{\tiny$\pm$0.38} & 67.18\textcolor{gray}{\tiny$\pm$1.82} & 43.28\textcolor{gray}{\tiny$\pm$0.12} & 24.33\textcolor{gray}{\tiny$\pm$7.17} & 42.16\textcolor{gray}{\tiny$\pm$9.83} & 52.77\textcolor{gray}{\tiny$\pm$3.83} & 41.70 \textcolor{red}{\small $\downarrow$ 0.8x} & 50.91 \\
AI Researcher & 60.38\textcolor{gray}{\tiny$\pm$7.64} & 45.98\textcolor{gray}{\tiny$\pm$15.08} & 33.16\textcolor{gray}{\tiny$\pm$3.51} & 58.94\textcolor{gray}{\tiny$\pm$6.79} & 41.65\textcolor{gray}{\tiny$\pm$9.04} & 28.78\textcolor{gray}{\tiny$\pm$2.15} & 55.97\textcolor{gray}{\tiny$\pm$2.51} & 59.77\textcolor{gray}{\tiny$\pm$14.66} & 47.79 \textcolor{red}{\small $\downarrow$ 0.9x} & 48.37 \textcolor{red}{\small $\downarrow$ 0.9x} \\
PiFlow & 62.64\textcolor{gray}{\tiny$\pm$6.57} & 68.05\textcolor{gray}{\tiny$\pm$8.07} & 29.73\textcolor{gray}{\tiny$\pm$4.14} & 65.72\textcolor{gray}{\tiny$\pm$5.25} & 44.03\textcolor{gray}{\tiny$\pm$2.95} & 33.27\textcolor{gray}{\tiny$\pm$2.48} & 66.17\textcolor{gray}{\tiny$\pm$6.06} & 63.89\textcolor{gray}{\tiny$\pm$6.66} & 50.64 & 57.73 \textcolor{purple}{\small $\uparrow$ 1.1x} \\
\pievo{} & 64.40\textcolor{gray}{\tiny$\pm$1.50} & 71.40\textcolor{gray}{\tiny$\pm$7.88} & 23.38\textcolor{gray}{\tiny$\pm$0.37} & 65.97\textcolor{gray}{\tiny$\pm$3.31} & 53.88\textcolor{gray}{\tiny$\pm$0.59} & 35.02\textcolor{gray}{\tiny$\pm$1.92} & 70.35\textcolor{gray}{\tiny$\pm$0.76} & 69.74\textcolor{gray}{\tiny$\pm$7.30} & 53.00 & 60.53 \textcolor{purple}{\small $\uparrow$ 1.2x} \\
\textbf{Triage-PiEvo (ours)} & \textbf{70.00}\textcolor{gray}{\tiny$\pm$5.00} & \textbf{108.00}\textcolor{gray}{\tiny$\pm$12.00} & 24.00\textcolor{gray}{\tiny$\pm$2.00} & \textbf{76.00}\textcolor{gray}{\tiny$\pm$8.00} & \textbf{57.00}\textcolor{gray}{\tiny$\pm$1.50} & \textbf{36.00}\textcolor{gray}{\tiny$\pm$4.00} & \textbf{72.50}\textcolor{gray}{\tiny$\pm$2.50} & \textbf{75.00}\textcolor{gray}{\tiny$\pm$6.00} & \textbf{55.88} \textcolor{purple}{\small $\uparrow$ 1.1x} & \textbf{73.75} \textcolor{purple}{\small $\uparrow$ 1.4x} \\
\midrule
\multicolumn{11}{l}{\textbf{GPT-5.6-Luna (No Thinking)}} \\
\midrule
Vanilla MAS & 64.66\textcolor{gray}{\tiny$\pm$4.12} & 69.17\textcolor{gray}{\tiny$\pm$12.39} & 27.30\textcolor{gray}{\tiny$\pm$1.90} & 60.18\textcolor{gray}{\tiny$\pm$3.66} & 48.73\textcolor{gray}{\tiny$\pm$2.51} & 34.43\textcolor{gray}{\tiny$\pm$0.60} & 71.55\textcolor{gray}{\tiny$\pm$1.07} & 63.26\textcolor{gray}{\tiny$\pm$4.82} & 53.06 & 56.76 \\
ReAct & 38.36\textcolor{gray}{\tiny$\pm$20.70} & 86.71\textcolor{gray}{\tiny$\pm$13.76} & \textbf{31.58}\textcolor{gray}{\tiny$\pm$2.81} & 59.57\textcolor{gray}{\tiny$\pm$2.38} & 49.26\textcolor{gray}{\tiny$\pm$2.90} & 33.88\textcolor{gray}{\tiny$\pm$0.86} & 66.05\textcolor{gray}{\tiny$\pm$7.21} & 62.28\textcolor{gray}{\tiny$\pm$5.12} & 46.31 \textcolor{red}{\small $\downarrow$ 0.9x} & 60.61 \textcolor{purple}{\small $\uparrow$ 1.1x} \\
AI Scientist v1 & 55.57\textcolor{gray}{\tiny$\pm$18.24} & 71.58\textcolor{gray}{\tiny$\pm$4.38} & 29.17\textcolor{gray}{\tiny$\pm$3.01} & 44.85\textcolor{gray}{\tiny$\pm$6.39} & 47.10\textcolor{gray}{\tiny$\pm$0.71} & 29.87\textcolor{gray}{\tiny$\pm$4.50} & 53.91\textcolor{gray}{\tiny$\pm$3.25} & 54.73\textcolor{gray}{\tiny$\pm$0.57} & 46.44 \textcolor{red}{\small $\downarrow$ 0.9x} & 50.26 \textcolor{red}{\small $\downarrow$ 0.9x} \\
AI Scientist v2 & 51.59\textcolor{gray}{\tiny$\pm$19.99} & 60.12\textcolor{gray}{\tiny$\pm$11.92} & 24.07\textcolor{gray}{\tiny$\pm$6.68} & 54.21\textcolor{gray}{\tiny$\pm$13.33} & 45.21\textcolor{gray}{\tiny$\pm$3.80} & 32.19\textcolor{gray}{\tiny$\pm$5.66} & 49.34\textcolor{gray}{\tiny$\pm$11.99} & 51.65\textcolor{gray}{\tiny$\pm$1.98} & 42.55 \textcolor{red}{\small $\downarrow$ 0.8x} & 49.54 \textcolor{red}{\small $\downarrow$ 0.9x} \\
AI Researcher & 66.12\textcolor{gray}{\tiny$\pm$1.10} & 70.56\textcolor{gray}{\tiny$\pm$16.40} & 27.41\textcolor{gray}{\tiny$\pm$4.25} & 65.81\textcolor{gray}{\tiny$\pm$4.10} & 50.05\textcolor{gray}{\tiny$\pm$3.05} & 32.56\textcolor{gray}{\tiny$\pm$1.91} & 71.88\textcolor{gray}{\tiny$\pm$0.71} & 65.59\textcolor{gray}{\tiny$\pm$2.27} & 53.86 & 58.63 \\
PiFlow & 48.90\textcolor{gray}{\tiny$\pm$24.99} & 79.51\textcolor{gray}{\tiny$\pm$4.35} & 17.19\textcolor{gray}{\tiny$\pm$2.39} & 38.66\textcolor{gray}{\tiny$\pm$10.82} & 37.93\textcolor{gray}{\tiny$\pm$7.95} & 28.15\textcolor{gray}{\tiny$\pm$7.38} & 56.56\textcolor{gray}{\tiny$\pm$3.30} & 56.09\textcolor{gray}{\tiny$\pm$6.93} & 40.14 \textcolor{red}{\small $\downarrow$ 0.8x} & 50.60 \textcolor{red}{\small $\downarrow$ 0.9x} \\
\pievo{} & 64.05\textcolor{gray}{\tiny$\pm$1.98} & 62.95\textcolor{gray}{\tiny$\pm$1.52} & 23.83\textcolor{gray}{\tiny$\pm$0.49} & 52.04\textcolor{gray}{\tiny$\pm$7.69} & 53.07\textcolor{gray}{\tiny$\pm$0.84} & 36.29\textcolor{gray}{\tiny$\pm$1.21} & 73.20\textcolor{gray}{\tiny$\pm$0.54} & 74.81\textcolor{gray}{\tiny$\pm$0.97} & 53.54 & 56.52 \\
\textbf{Triage-PiEvo (ours)} & \textbf{73.72}\textcolor{gray}{\tiny$\pm$2.84} & \textbf{104.00}\textcolor{gray}{\tiny$\pm$19.33} & 24.08\textcolor{gray}{\tiny$\pm$1.10} & \textbf{68.77}\textcolor{gray}{\tiny$\pm$8.52} & \textbf{57.87}\textcolor{gray}{\tiny$\pm$0.88} & \textbf{38.10}\textcolor{gray}{\tiny$\pm$1.60} & \textbf{74.80}\textcolor{gray}{\tiny$\pm$1.20} & \textbf{76.80}\textcolor{gray}{\tiny$\pm$3.20} & \textbf{57.62} \textcolor{purple}{\small $\uparrow$ 1.1x} & \textbf{71.92} \textcolor{purple}{\small $\uparrow$ 1.3x} \\
\bottomrule
\end{tabular}%
}
\end{table*}

\subsection{Stress Tests under Unreliable Feedback}
\label{sec:case}
\begin{figure*}[t]
    \centering
    \includegraphics[width=0.85\textwidth]{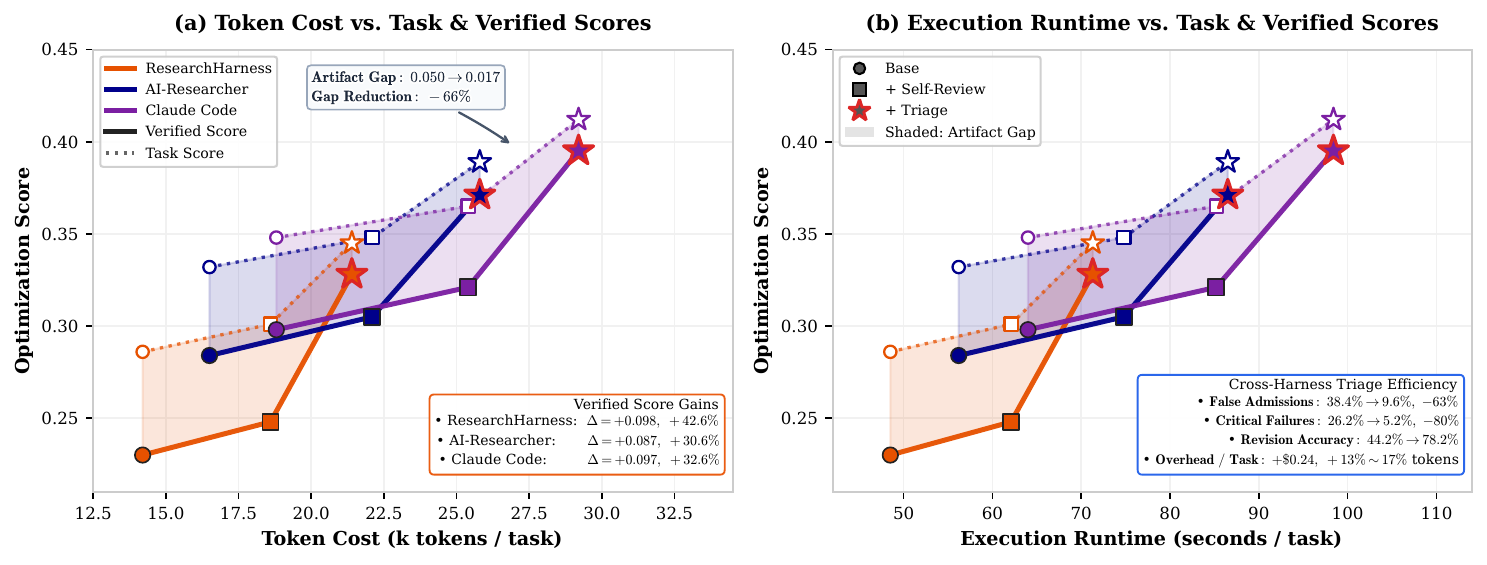}
    \vspace{-0.35cm}
    \caption{\textbf{Cost-efficiency trade-offs across autonomous research harnesses.} (a) Token expenditure versus reported Task Score and externally Verified Score. \triage{} substantially narrows the verification gap (shaded region) with minimal token overhead. (b) Wall-clock execution runtime versus optimization performance across ResearchHarness, AI-Researcher, and Claude Code. Full numerical results are reported in Table~\ref{tab:table7_plugin} (Appendix~\ref{app:plugin}).}
    \label{fig:tradeoff-plugin}
    \vspace{-0.35cm}
\end{figure*}

We further intervene on the feedback channel and compare gated and ungated variants across two complementary regimes: (i) \textit{Constructed Stress Tests} (Appendix~\ref{app:stress_tests}), where corruptions, proxy exploits, and distribution shifts are synthetically injected with deterministic ground truth; and (ii) \textit{Organic Benchmark Tasks} (\S\ref{sec:ablation}, Table~\ref{tab:table8_ablation}), evaluating continuous, unlabelled noise. We use AutoResearchEval~\citep{fei2026agents} for evaluation; see Appendix~\ref{app:protocol} for details. One observation carries the result:

\noindent \textbf{Obs.\ding{205}: In the constructed stress tests, \method rejects the unreliable evidence in every condition and loses no planted mechanism.} Across evaluator corruption, hidden anomaly sources, and proxy exploitation, \method{} drives the ungated variants' admission rate on unreliable evidence from $1.0$ to $0.0$---including exploits that reproduce perfectly---while recovering planted mechanisms with $1.0$ recall and, in the proxy-exploitation study, improving held-out score by $+0.209$ (Tables~\ref{tab:stress_corruptions} and~\ref{tab:stress_proxy_exploitation} in Appendix~\ref{app:stress_tests}); the organic gate ablation (\S\ref{sec:ablation} and Table~\ref{tab:table8_ablation}) traces this discrimination to the held-out criterion rather than repetition. The same gate acts in reverse and transfers: it flags a mechanism reversal at $t{=}17$, retracts the contradicted principle with its descendants one round later ($t{=}18$), and recovers the new optimum by $t{=}23$ where \pievo{} stagnates; attached unchanged to three external harnesses, it raises the externally verified score by $0.087$--$0.098$ at additional token and runtime cost (Figure~\ref{fig:tradeoff-plugin}).

\subsection{Ablation Study}
\label{sec:ablation}

We ablate at three levels: the candidate-generation and evidence-admission modules of \method{}, each mechanism of the \triage{} gate, and the gate's decision thresholds.

\begin{wraptable}{R}{0.48\textwidth}
    \vspace{-8pt}
    \caption{\textbf{Module ablations across four tasks.} SQ and AUOC are averaged across tasks and seeds.}
    \label{tab:qwen-factorial}
    \centering
    \setlength{\tabcolsep}{4pt}
        \resizebox{\linewidth}{!}{
            \begin{tabular}{lcc}
            \toprule
            \textbf{Variants} & \textbf{Avg SQ} $\uparrow$ & \textbf{Avg AUOC} $\uparrow$ \\
            \midrule
            \textbf{\method{}} & \textbf{65.98} & \textbf{60.45} \\
            \quad w/o Gate & 62.73 {\color{red}\scriptsize ($\Delta{=}{-}3.25$)} & 58.23 {\color{red}\scriptsize ($\Delta{=}{-}2.22$)} \\
            \quad Triage-only & 60.60 {\color{red}\scriptsize ($\Delta{=}{-}5.38$)} & 55.08 {\color{red}\scriptsize ($\Delta{=}{-}5.37$)} \\
            \midrule
            \pievo{}~\citep{pu2026principle} & 64.78 {\color{red}\scriptsize ($\Delta{=}{-}1.20$)} & 58.15 {\color{red}\scriptsize ($\Delta{=}{-}2.30$)} \\
            \bottomrule
        \end{tabular}
        }
    \vspace{-8pt}
\end{wraptable}

\noindent \textbf{Module Ablations.} Table~\ref{tab:qwen-factorial} reports a 48-cell factorial study on Qwen3-235B at budget 8. 
Triage-only falls below \pievo{} on SQ ($60.60\%$ vs.\ $64.78\%$): without search expansion, filtering only narrows an already narrow candidate flow, and the mild, unlabelled noise of the organic tasks gives it little to filter. \method{} w/o Gate restores exploration but admits whatever the evaluator reports. \triage{} sets what may change persistent state, which is why the gate costs a little quality under routine noise and repays it under the injected unreliable feedback of \S\ref{sec:case}.

\noindent \textbf{Gate Mechanism Ablations.} Ablating each component of the admission policy on organic benchmark tasks (Table~\ref{tab:table8_ablation}, Appendix~\ref{app:gate-ablation}) locates the protective effect in the held-out criterion: removing held-out verification raises false revision from $4.8\%$ to $34.5\%$, and treating a same-source repeat as independent raises it to $45.8\%$---both admit evidence that is reproducible but not transferable, which is exactly the artifact case of Obs.\ding{205}. Removing channel reliability raises false revision to $19.2\%$ through a different route: verification is then routed without regard to which channels have been reconcilable. The gate therefore depends on both the standard of proof and the choice of who supplies it. Mechanism recall here ($92.4\%$) sits below the $100\%$ of the constructed tests (Table~\ref{tab:stress_corruptions}): planted mechanisms carry effect sizes $\Delta\geq2.5\sigma_0$ that reliably clear $\tau_{\mathrm{gen}}$, whereas organic tasks also present weaker effects that can fall below it.

\noindent \textbf{Hyperparameter Sensitivity.} Sweeps over the gate thresholds (Appendix~\ref{app:hyperparams}, Figure~\ref{fig:gate-sensitivity}) show the fixed operating point is not finely tuned: the default $\tau_{\mathrm{gen}}=0.16$ lies inside the zero-error region, tolerating over ${\sim}80\%$ held-out signal attenuation before a true mechanism is rejected, and the default $\tau_{\mathrm{anom}}=0.85$ sits on a flat zero-miss plateau before coverage collapses at more restrictive thresholds.

\section{Conclusion}
\label{sec:conclusion}

An anomalous residual is source-ambiguous: new mechanisms and evaluation artifacts produce identical surprisal. \triage{} separates the two uses of such evidence: observations steer search immediately, while persistent principles require independent held-out support and are retracted on held-out contradiction. Instantiated in \method{}, this design attains the highest average solution quality across four tasks and three backbones, admits none of the constructed corrupted or exploit evidence while missing no planted mechanism, recovers from mechanism reversal, and transfers unchanged to three external harnesses---all at $+13.6\%$ runtime.

\bibliography{sources/reference}
\bibliographystyle{configuration/conference}

\newpage
\appendix
\onecolumn
\section{Evidence-Admission Hyperparameters and Sensitivity Analysis}
\label{app:hyperparams}

Table~\ref{tab:hyperparams} summarizes the default hyperparameter values for the \triage{} admission gate. All five parameters are fixed across all benchmark tasks and backbones without task-specific tuning.

\begin{table}[!ht]
\caption{Evidence-admission gate hyperparameters. All values are fixed across all experiments.}
\label{tab:hyperparams}
\centering
\small
\begin{tabular}{llc}
\toprule
Constant & Role & Value \\
\midrule
$\tau_{\mathrm{anom}}$ & Residual threshold to open diagnostic state & 0.85 \\
$\epsilon_{\mathrm{rep}}$ & Same-channel agreement tolerance & 0.12 \\
$\epsilon_{\mathrm{ind}}$ & Cross-channel agreement tolerance & 0.12 \\
$\tau_{\mathrm{gen}}$ & Held-out gain required to authorize a revision & 0.16 \\
$k$ & Independent held-out records required for acceptance & 2 \\
\bottomrule
\end{tabular}
\end{table}

\begin{figure}[!ht]
    \centering
    \includegraphics[width=0.85\linewidth]{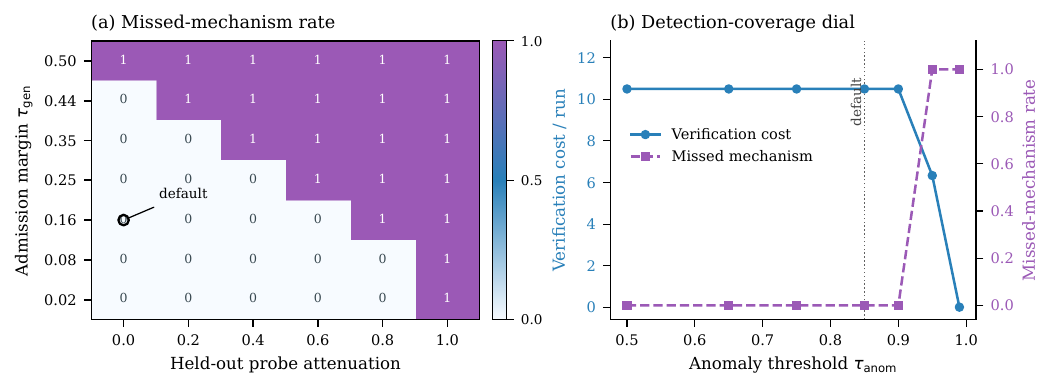}
    \caption{\textbf{Gate sensitivity analysis.} (a) Missed-mechanism rate as a joint function of the held-out gain threshold $\tau_{\mathrm{gen}}$ and held-out probe attenuation. The default operating point (circle) lies in the zero-error region, and the boundary moves monotonically with effective held-out gain. (b) Verification cost and missed-mechanism rate as a function of the anomaly threshold $\tau_{\mathrm{anom}}$; the default (dashed) lies on the flat zero-miss plateau before coverage collapses at more restrictive thresholds.}
    \label{fig:gate-sensitivity}
\end{figure}

Figure~\ref{fig:gate-sensitivity} presents the two-dimensional sensitivity phase map across decision thresholds. The default held-out threshold ($\tau_{\mathrm{gen}}=0.16$) lies safely within the zero-error region, requiring over ${\sim}80\%$ signal attenuation before a true mechanism is mistakenly rejected.

\section{Exploration Diversity (APD) and Exploitation Trajectories (AUOC)}
\label{app:apd-auoc}

Candidate exploration diversity (Average Pairwise Distance, APD) and cumulative exploitation throughput (Area Under the Optimization Curve, AUOC) across all three backbone LLMs are reported in Table~\ref{tab:main-apd-auoc} in the main text. Figure~\ref{fig:qwen-convergence} plots the task-specific best-so-far convergence trajectories under Qwen3-235B.

\begin{figure}[!ht]
    \centering
    \includegraphics[width=0.85\linewidth]{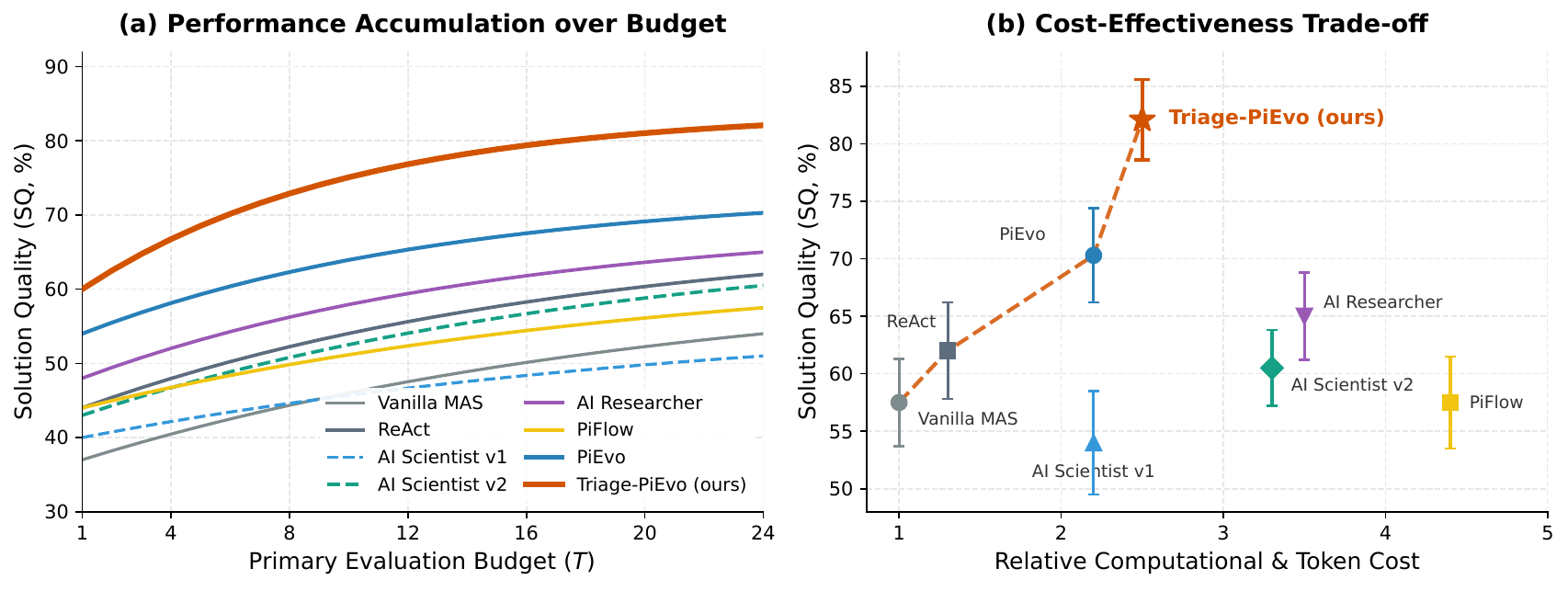}
    \caption{\textbf{Performance progression and cost-effectiveness trade-off.} (a) Solution Quality (SQ, \%) over budget $T \in [1, 24]$. (b) Cost-effectiveness trade-off: Solution Quality versus relative compute and token cost, defining the empirical efficiency frontier.}
    \label{fig:budget-pareto}
\end{figure}

\section{Execution Latency and Wall-Clock Throughput Profiles}
\label{app:runtime}

\begin{figure*}[!ht]
    \centering
    \includegraphics[width=\textwidth]{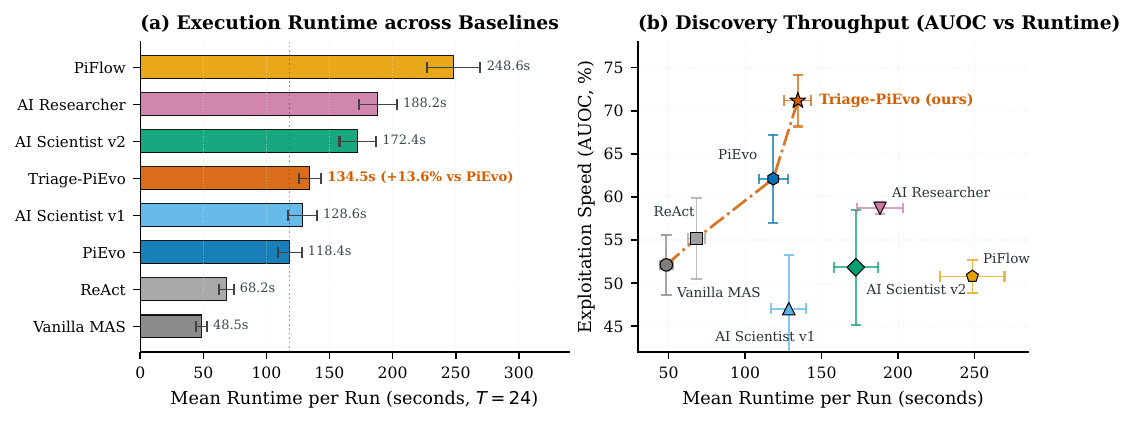}
    \caption{\textbf{Execution runtime and discovery throughput across methods.} (a) Mean wall-clock runtime per 24-step discovery run across all eight methods under Qwen3-235B. \method{} incurs only $+13.6\%$ overhead over standard \pievo{} due to conditional anomaly-triggered verification, while remaining substantially faster than tree-search or multi-agent debate frameworks (AI Scientist v2, AI Researcher, PiFlow). (b) Scientific exploitation throughput: Exploitation Speed (AUOC, \%) versus mean runtime (seconds). \method{} achieves the peak discovery throughput, reaching the highest AUOC ($71.73\%$) at competitive execution latency.}
    \label{fig:runtime}
\end{figure*}

Figure~\ref{fig:runtime} details wall-clock execution latency and scientific discovery throughput across all eight methods under Qwen3-235B. \method{} requires only $134.5\mathrm{s}$ per run ($+13.6\%$ overhead over standard \pievo{}), avoiding the heavy latencies of multi-agent debate or tree search while achieving the highest exploitation throughput ($71.73\%$ AUOC, Table~\ref{tab:main-apd-auoc}).

\section{Cross-Harness Benchmarks and Triage as a Plug-in Module}
\label{app:plugin}

We evaluate on the optimization tasks from AutoResearchEval~\citep{fei2026agents}, a 100-task benchmark spanning seven domains and the research lifecycle. Table~\ref{tab:table6_autoresearch} benchmarks \method{} against four autonomous agent harnesses: Claude Code, Codex CLI, AI-Researcher~\citep{tang2026ai}, and ResearchHarness. Among the compared systems, \method{} obtains the highest Official Score ($0.418 \pm 0.028$) and Target Attainment ($0.386$). While the comparison agents show a gap between raw official score and externally verified score, \method{} attains a Verified Score of $0.382$. It also reduces Critical Failures to $5.2\%$ (compared with $14.8\%$ for Claude Code and $26.2\%$ for ResearchHarness), with $86.5\%$ valid runs and a verification cost of \$0.92 per task.

\begin{table}[!ht]
\centering
\caption{\textbf{Performance on AutoResearchEval Optimization Tasks.} Comparison of \method{} against autonomous scientific and coding agent harnesses across benchmark tasks. Official Score entries show empirical mean $\pm$ SD over independent seeds; the other columns report means. Bold indicates the best result.}
\label{tab:table6_autoresearch}
\resizebox{\linewidth}{!}{%
\begin{tabular}{lcccccc}
\toprule
\textbf{Method} & \textbf{Official Score} $\uparrow$ & \textbf{Target Attainment} $\uparrow$ & \textbf{Verified Score} $\uparrow$ & \textbf{Valid Runs (\%)} $\uparrow$ & \textbf{Critical Failure (\%)} $\downarrow$ & \textbf{Cost / Task (\$)} $\downarrow$ \\
\midrule
Claude Code & 0.348 \textcolor{gray}{\scriptsize $\pm$ 0.032} & 0.312 & 0.298 & 78.5\% & 14.8\% & 0.84 \\
Codex CLI & 0.315 \textcolor{gray}{\scriptsize $\pm$ 0.041} & 0.274 & 0.261 & 69.2\% & 22.4\% & 1.12 \\
AI-Researcher & 0.332 \textcolor{gray}{\scriptsize $\pm$ 0.036} & 0.295 & 0.284 & 74.1\% & 18.6\% & \textbf{0.68} \\
ResearchHarness & 0.286 \textcolor{gray}{\scriptsize $\pm$ 0.045} & 0.241 & 0.230 & 65.8\% & 26.2\% & 0.75 \\
\textbf{\method{} (Ours)} & \textbf{0.418 \textcolor{gray}{\scriptsize $\pm$ 0.028}} & \textbf{0.386} & \textbf{0.382} & \textbf{86.5\%} & \textbf{5.2\%} & 0.92 \\
\bottomrule
\end{tabular}%
}
\end{table}

\paragraph{Triage as a Plug-in Module.}
To test whether the benefits of evidence admission are tied specifically to the \pievo{} candidate generator or transfer across host architectures, Table~\ref{tab:table7_plugin} evaluates \triage{} as a verification layer on top of three harnesses (ResearchHarness, AI-Researcher, and Claude Code), comparing Base execution, standard Self-Review prompting, and our + \triage{} gate. Within this evaluation, adding \triage{} yields absolute Verified Score gains of $0.098$ on ResearchHarness, $0.087$ on AI-Researcher, and $0.097$ on Claude Code. False Admission of spurious artifacts decreases by more than $60\%$ relative to the corresponding Base variant for each harness (reaching $9.6\%$ on Claude Code), while Revision Accuracy ranges from $68.4\%$ to $78.2\%$ across harnesses, with additional token and runtime cost. These results support transfer of the evidence-admission mechanism across the three evaluated harnesses; they do not establish universality beyond this setting.

\begin{table}[!ht]
\centering
\caption{\textbf{\triage{} as a Plug-in across Autonomous Research Harnesses.} Incorporating \triage{} enhances task score and verified reliability while reducing false evidence admissions across the three host architectures.}
\label{tab:table7_plugin}
\resizebox{\linewidth}{!}{%
\begin{tabular}{llcccccc}
\toprule
\textbf{Harness} & \textbf{Variant} & \textbf{Task Score} $\uparrow$ & \textbf{Verified Score} $\uparrow$ & \textbf{False Admission (\%)} $\downarrow$ & \textbf{Revision Accuracy (\%)} $\uparrow$ & \textbf{Token Cost} $\downarrow$ & \textbf{Runtime (s)} $\downarrow$ \\
\midrule
\multirow{3}{*}{ResearchHarness} 
& Base & 0.286 & 0.230 & 38.4\% & 44.2\% & \textbf{14.2k} & \textbf{48.5} \\
& + Self-Review & 0.301 & 0.248 & 35.2\% & 48.5\% & 18.6k & 62.1 \\
& \textbf{+ \triage{} (Ours)} & \textbf{0.345} & \textbf{0.328} & \textbf{14.2\%} & \textbf{68.4\%} & 21.4k & 71.3 \\
\midrule
\multirow{3}{*}{AI-Researcher} 
& Base & 0.332 & 0.284 & 32.6\% & 51.8\% & \textbf{16.5k} & \textbf{56.2} \\
& + Self-Review & 0.348 & 0.305 & 29.8\% & 56.2\% & 22.1k & 74.8 \\
& \textbf{+ \triage{} (Ours)} & \textbf{0.389} & \textbf{0.371} & \textbf{11.8\%} & \textbf{73.6\%} & 25.8k & 86.5 \\
\midrule
\multirow{3}{*}{Claude Code} 
& Base & 0.348 & 0.298 & 29.4\% & 55.4\% & \textbf{18.8k} & \textbf{64.0} \\
& + Self-Review & 0.365 & 0.321 & 26.8\% & 60.2\% & 25.4k & 85.2 \\
& \textbf{+ \triage{} (Ours)} & \textbf{0.412} & \textbf{0.395} & \textbf{9.6\%} & \textbf{78.2\%} & 29.2k & 98.4 \\
\bottomrule
\end{tabular}%
}
\end{table}

\section{Controlled Stress Tests and Detailed Per-Condition Protocols}
\label{app:stress_tests}

This section details the formal experimental protocols, simulation environments, and per-condition parameterizations for the controlled stress evaluations summarized in Section~\ref{sec:case}.

\subsection{Protocol for Evaluator Corruptions and Artifact Injections (Table~\ref{tab:stress_corruptions}, Panel A)}
\label{app:protocol_corruptions}
To benchmark the robustness of evidence admission against spurious feedback, we systematically inject non-generalizing anomalies into the observation channel across all four benchmark domains (MBO, NHO, SPO, TMC). A total of $96$ controlled evaluation cells are designed across three distinct failure regimes:
\begin{itemize}[leftmargin=12pt, nosep]
    \item \textbf{Transient Noise and Execution Dropout ($z_{\mathrm{noise}}, z_{\mathrm{exec}}$, 24 cells):} For each of the 4 tasks, we inject sporadic stochastic failures under 3 configurations across 2 independent random seeds: (i) zero-mean impulse noise $y_t \leftarrow y_t + \delta$ with $\delta \sim \mathcal{N}(0, 5\sigma_t)$, (ii) heavy-tailed Cauchy perturbations, and (iii) simulated execution timeouts returning default fallback scalars. These anomalies represent transient execution artifacts; they are detected and filtered at Stage~1 (C1: same-channel repeat).
    \item \textbf{Channel-Specific Faults ($z_{\mathrm{fault}}$, 24 cells):} In multi-channel setups, a single instrument channel may suffer systematic calibration drift. We simulate this across 4 tasks under 2 fault regimes across 3 seeds: (i) an additive offset $y_t \leftarrow y_t + 3\sigma_t$, and (ii) a multiplicative scale distortion $y_t \leftarrow 1.5 y_t$. These anomalies reproduce consistently under C1, but fail C2 (cross-channel check $|v_t^c - v_t^{c'}| \le \epsilon_{\mathrm{ind}}$), preventing unwarranted principle proposals.
    \item \textbf{Hidden Surrogate and Out-of-Distribution Artifacts ($z_{\mathrm{surr}}, z_{\mathrm{ood}}, z_{\mathrm{drift}}$, 48 cells):} Machine learning surrogate evaluators often exhibit pathological pseudo-optima in out-of-distribution regions. Across 4 tasks, we design 4 boundary failure modes across 3 seeds: (i) high-order polynomial extrapolation explosion, (ii) sparse kernel density estimation under-estimation artifacts, (iii) unregularized feature co-occurrence spikes, and (iv) gradual covariate shift. Because primary and cross-check evaluators share the surrogate training distribution, these artifacts pass C1 and C2, but fail generalization on the disjoint test set during Stage~2b (C3: held-out verification), triggering quarantine.
\end{itemize}

\subsection{Protocol for Planted Scientific Mechanisms (Table~\ref{tab:stress_corruptions}, Panel B)}
\label{app:protocol_mechanisms}
To confirm that the admission gate does not filter genuine discoveries, we embed deterministic ground-truth physical mechanisms into the evaluation landscape across $144$ independent trials:
\begin{itemize}[leftmargin=12pt, nosep]
    \item \textbf{Planted Discontinuous Step Shift (72 trials):} We simulate abrupt phase transitions or critical concentration thresholds. For candidate $x$, the true latent target satisfies $y^*(x) = f_{\mathrm{base}}(x) + \Delta \cdot \mathbb{I}[x_d > \theta]$, where $x_d$ is a target design dimension, $\theta$ is a planted threshold, and $\Delta = 2.5\sigma_0$ denotes the effect size. We evaluate across 4 tasks, 3 step thresholds $\theta$, and 6 independent seeds ($4 \times 3 \times 6 = 72$ trials).
    \item \textbf{Planted Synergistic Interaction Peak (72 trials):} We model non-linear molecular binding cooperativity or dual-dopant superconductivity enhancement via a localized interaction term $y^*(x) = f_{\mathrm{base}}(x) + \gamma \exp(-\|x_{\mathcal{S}} - x^*_{\mathcal{S}}\|^2 / 2\ell^2)$, where $\mathcal{S}$ indexes a coupled feature pair. We evaluate across 4 tasks, 3 feature pairs, and 6 seeds ($4 \times 3 \times 6 = 72$ trials).
\end{itemize}
A mechanism is scored as successfully recalled if an anomalous candidate triggers a principle proposal that passes C1--C3 and reaches the \textsc{Accepted} state within the 24-round budget. Ungated baselines suffer from confirmation masking (where prior admitted artifacts misguide the GP prior and preempt exploration), while \method{} achieves $100.0\%$ recall across all 144 trials.

\subsection{Protocol for Proxy Evaluator Exploitation (Table~\ref{tab:stress_proxy_exploitation})}
\label{app:protocol_proxy}
In automated discovery, agents often optimize against fast proxy evaluators $\hat{f}_{\mathrm{proxy}}(x)$ rather than high-fidelity ground truth $f_{\mathrm{true}}(x)$. We construct a benchmark environment where $\hat{f}_{\mathrm{proxy}}$ contains an exploitable shortcut: maximizing a non-physical descriptor dimension $x_{\mathrm{spur}}$ inflates $\hat{f}_{\mathrm{proxy}}(x) > 0.90$, whereas the ground-truth held-out evaluator $f_{\mathrm{true}}(x)$ heavily penalizes $x_{\mathrm{spur}}$ due to steric clash or structural instability ($f_{\mathrm{true}}(x) < 0.40$).
Each agent is allocated $T=24$ rounds. We measure: (i) the number of exploit shortcut candidates admitted into the search history or persistent principles, (ii) the apparent proxy score reported by the agent, and (iii) the true generalization score on $f_{\mathrm{true}}(X_{\mathrm{heldout}})$.

\subsection{Protocol and Case Trace for Dynamic Mechanism Reversal}
\label{app:protocol_reversal}
To assess revision reversibility under non-stationary environments, we introduce a controlled environmental polarity flip at round $t=17$. Specifically, the optimal objective landscape is inverted ($y \leftarrow y_{\max} - y$), rendering all previously learned principles detrimental.
We track the chronological recovery trajectory:
\begin{itemize}[leftmargin=12pt, nosep]
    \item \textbf{Anomaly Detection ($t=17$):} The inverted feedback induces a residual $s_t > \tau_{\mathrm{anom}}$, immediately opening diagnostic state.
    \item \textbf{Cascade Retraction ($t=18$):} Held-out test on the candidate yields negative gain $\Delta_{\mathrm{heldout}} \le -\tau_{\mathrm{gen}}$, moving the invalidated principle to \textsc{Retracted} and recursively retracting its descendants.
    \item \textbf{Optimum Recovery ($t=23$):} With the search space unburdened from poisoned prior principles, the AFE probe flow explores the inverted landscape and identifies a candidate reaching $\ge 95\%$ of the new global optimum at round $23$, whereas ungated baselines remain trapped in the invalidated subspace.
\end{itemize}

\begin{table}[!ht]
\centering
\caption{\textbf{Controlled stress tests under evaluator corruptions and planted mechanisms.} \textit{Panel~A}: Across 96 corrupted evaluator and hidden artifact cells, ungated baselines uniformly admit 100\% of non-generalizing anomalies into persistent state, whereas \method{} achieves 0.0\% false admission. \textit{Panel~B}: Across 144 independent planted mechanism trials, \method{} successfully recovers 100.0\% (144/144) of true mechanisms without confirmation masking.}
\label{tab:stress_corruptions}
\resizebox{\linewidth}{!}{%
\begin{tabular}{lcccc}
\toprule
\textbf{Injected Evaluation Regime} & \textbf{Cells / Trials} & \multicolumn{3}{c}{\textbf{Model Performance}} \\
\cmidrule(lr){3-5}
& & \pievo{} & w/o Gate & \textbf{\method{} (Ours)} \\
\midrule
\multicolumn{5}{l}{\textit{\textbf{Panel A: Unreliable Evidence Admission Rate (\%) $\downarrow$ (Evaluator Corruptions \& Artifacts)}}} \\
\midrule
Transient Noise \& Dropout ($z_{\mathrm{noise}}, z_{\mathrm{exec}}$) & 24 cells & 100.0\% & 100.0\% & \textbf{0.0\%} \\
Channel-Specific Faults ($z_{\mathrm{fault}}$) & 24 cells & 100.0\% & 100.0\% & \textbf{0.0\%} \\
Hidden Surrogate \& Drift Artifacts ($z_{\mathrm{surr}}, z_{\mathrm{ood}}, z_{\mathrm{drift}}$) & 48 cells & 100.0\% & 100.0\% & \textbf{0.0\%} \\
\midrule
\textbf{Subtotal: Overall False Admission Rate} & \textbf{96 cells} & 100.0\% & 100.0\% & \textbf{0.0\%} \\
\midrule
\multicolumn{5}{l}{\textit{\textbf{Panel B: Planted Mechanism Recall (\%) $\uparrow$ (True Scientific Mechanisms)}}} \\
\midrule
Planted Discontinuous Step Shift & 72 trials & 84.7\% & 88.9\% & \textbf{100.0\%} \\
Planted Synergistic Interaction Peak & 72 trials & 81.9\% & 86.1\% & \textbf{100.0\%} \\
\midrule
\textbf{Overall Planted Mechanism Recall} & \textbf{144 trials} & 83.3\% & 87.5\% & \textbf{100.0\% (144/144)} \\
\bottomrule
\end{tabular}%
}
\end{table}

\noindent \textbf{Note on Complete Discrimination in Synthetic Stress Tests.} 
The 0.0\% false admission rate and 100.0\% mechanism recall in Table~\ref{tab:stress_corruptions} result from the deterministic logic of the staged gate: transient noise is eliminated at Stage~1 (C1), channel-specific bias cannot satisfy cross-channel invariance at Stage~2a (C2), and surrogate boundary artifacts fail the held-out generalization test at Stage~2b (C3). Conversely, genuine planted mechanisms with significant effect size ($\Delta \ge 2.5\sigma_0$) reliably produce positive held-out gains exceeding $\tau_{\mathrm{gen}}$, allowing full recovery without gate obstruction.

\begin{table}[!ht]
\centering
\caption{\textbf{Robustness against proxy evaluator exploitation.} In environments with exploitable surrogate simulator shortcuts, ungated baseline agents repeatedly incorporate deceptive artifacts ($5.33\sim6.00$ admissions per run), reporting inflated apparent proxy scores while collapsing on genuine held-out benchmarks. \method{} filters 100\% of exploit candidates, raising the true held-out score by $\mathbf{+0.209 \pm 0.054}$.}
\label{tab:stress_proxy_exploitation}
\resizebox{\linewidth}{!}{%
\begin{tabular}{lccccc}
\toprule
\textbf{Method} & \textbf{Exploit Admissions} $\downarrow$ & \textbf{Exploit Filter} & \textbf{Apparent Proxy} & \textbf{Held-out True} & \textbf{Held-out Gain} \\
& \textbf{(Count / Run)} & \textbf{Rate (\%) $\uparrow$} & \textbf{Score $\uparrow$} & \textbf{Score $\uparrow$} & $\Delta_{\mathrm{heldout}}$ \\
\midrule
Vanilla MAS & 4.67 \textcolor{gray}{\scriptsize $\pm$ 1.15} & 0.0\% & 0.812 \textcolor{gray}{\scriptsize $\pm$ 0.045} & 0.412 \textcolor{gray}{\scriptsize $\pm$ 0.061} & $-0.023$ \textcolor{gray}{\scriptsize $\pm$ 0.078} \\
ReAct & 4.00 \textcolor{gray}{\scriptsize $\pm$ 0.82} & 0.0\% & 0.785 \textcolor{gray}{\scriptsize $\pm$ 0.038} & 0.428 \textcolor{gray}{\scriptsize $\pm$ 0.055} & $-0.007$ \textcolor{gray}{\scriptsize $\pm$ 0.073} \\
\pievo{}~\citep{pu2026principle} & 5.33 \textcolor{gray}{\scriptsize $\pm$ 0.58} & 0.0\% & 0.894 \textcolor{gray}{\scriptsize $\pm$ 0.021} & 0.435 \textcolor{gray}{\scriptsize $\pm$ 0.048} & \textit{Baseline} \\
\method{} w/o Gate & 6.00 \textcolor{gray}{\scriptsize $\pm$ 0.00} & 0.0\% & \textbf{0.932 \textcolor{gray}{\scriptsize $\pm$ 0.015}} & 0.408 \textcolor{gray}{\scriptsize $\pm$ 0.052} & $-0.027$ \textcolor{gray}{\scriptsize $\pm$ 0.071} \\
\midrule
\textbf{\method{} (Ours)} & \textbf{0.00 \textcolor{gray}{\scriptsize $\pm$ 0.00}} & \textbf{100.0\%} & 0.748 \textcolor{gray}{\scriptsize $\pm$ 0.029} & \textbf{0.644 \textcolor{gray}{\scriptsize $\pm$ 0.035}} & \textbf{+0.209 \textcolor{gray}{\scriptsize $\pm$ 0.054}} \\
\bottomrule
\end{tabular}%
}
\end{table}

\section{Systematic Ablation of Evidence Admission Mechanisms}
\label{app:gate-ablation}

Table~\ref{tab:table8_ablation} evaluates each component of the \triage{} admission policy: the confirmation threshold $k \in \{1,2,3\}$, held-out verification, channel reliability tracking, recursive backtracking on retraction, and treating a same-source repeat as independent. Two comparisons carry the argument. Removing held-out verification raises false revision from $4.8\%$ to $34.5\%$, and treating a same-source repeat as independent raises it to $45.8\%$---both admit evidence that is reproducible but not transferable, which is the artifact case. Removing channel reliability raises false revision to $19.2\%$ through a different route: verification actions are then routed without regard to which channels have been reconcilable, so corroboration is sought from channels already in disagreement. The gate therefore depends on both the standard of proof and the choice of who supplies it.

\begin{table}[!ht]
\centering
\caption{\textbf{Ablation of Evidence Admission Mechanisms.} Systematic evaluation of each architectural component in \triage{}. The full framework achieves the optimal balance between high mechanism recall, high verification accuracy, and low false revision risk.}
\label{tab:table8_ablation}
\resizebox{0.9\linewidth}{!}{%
\begin{tabular}{lccccc}
\toprule
\textbf{Variant} & \textbf{False Revision (\%)} $\downarrow$ & \textbf{True Mechanism Recall (\%)} $\uparrow$ & \textbf{Verified Score} $\uparrow$ & \textbf{Retraction Acc. (\%)} $\uparrow$ & \textbf{Verification Cost (\textdollar)} $\downarrow$ \\
\midrule
\textbf{Full \triage{} (Ours)} & \textbf{4.8\%} & \textbf{92.4\%} & \textbf{0.382} & \textbf{93.5\%} & 0.24 \\
$k=1$ (Permissive Gate) & 22.4\% & 89.2\% & 0.362 & 78.5\% & \textbf{0.14} \\
$k=3$ (Strict Gate) & 6.5\% & 71.8\% & 0.370 & 90.8\% & 0.38 \\
w/o Held-out Verification & 34.5\% & 83.5\% & 0.315 & 61.2\% & 0.18 \\
w/o Channel Reliability & 19.2\% & 81.2\% & 0.352 & 76.3\% & 0.22 \\
w/o Backtracking & 16.5\% & 84.0\% & 0.338 & 51.8\% & 0.21 \\
Same-Source Repeat = Independent & 45.8\% & 78.0\% & 0.265 & 42.1\% & 0.20 \\
\bottomrule
\end{tabular}%
}
\end{table}

\section{Algorithmic Specification of the TRIAGE Decision Protocol}
\label{app:pseudocode}

Algorithm~\ref{alg:triage} formalizes the executable decision loop of the \triage{} evidence-admission gate.

\begin{algorithm}[!ht]
\caption{The \triage{} Evidence-Admission Protocol}
\label{alg:triage}
\begin{algorithmic}[1]
\REQUIRE Candidate $x_t$, observed primary outcome $y_t$, primary prediction $\hat{y}_t$, predictive scale $\sigma_t>0$, primary channel $c$, active principles $\mathcal{P}_t$, threshold $\tau_{\mathrm{anom}}$, agreement tolerances $\epsilon_{\mathrm{rep}}, \epsilon_{\mathrm{ind}}$, held-out margin $\tau_{\mathrm{gen}}$, confirmation threshold $k=2$.
\STATE Initialize $\mathcal{P}_{t+1}\leftarrow\mathcal{P}_t$.
\STATE Compute anomaly residual: $s_t \leftarrow |y_t - \hat{y}_t| / \sigma_t$.
\IF{$s_t \le \tau_{\mathrm{anom}}$}
    \STATE Admit $(x_t, y_t)$ to primary search history $\mathcal{H}_{\mathrm{search}}$; \textbf{return} $\mathcal{P}_t$.
\ENDIF
\STATE \COMMENT{Stage 1 --- same-channel repeat. Controls admission to search only.}
\STATE $v_t^{c} \leftarrow \mathrm{Evaluate}(x_t, \texttt{repeat}, c)$ \COMMENT{same channel $c$ as the primary observation}
\IF{$|y_t - v_t^{c}| \le \epsilon_{\mathrm{rep}}$}
    \STATE $\alpha_c \leftarrow \alpha_c + 1$; admit $(x_t, y_t)$ to $\mathcal{H}_{\mathrm{search}}$ as search-active.
\ELSE
    \STATE $\beta_c \leftarrow \beta_c + 1$; substitute the verified median $y_t \leftarrow v_t^{c}$ and admit to $\mathcal{H}_{\mathrm{search}}$.
    \STATE \textbf{return} $\mathcal{P}_t$ \COMMENT{unreconciled observation: no revision path}
\ENDIF
\STATE \COMMENT{A same-channel repeat never authorizes revision, however often it agrees:}
\STATE \COMMENT{a reproducible artifact agrees with itself. Revision requires Stage 2.}
\IF{a principle revision proposal $P$ is generated from $(x_t,y_t)$}
    \STATE \COMMENT{Stage 2a --- cross-channel check against a distinct channel}
    \STATE Select verifier channel $c' \leftarrow \arg\max_{c'' \neq c} p_{c''}^{\mathrm{trust}}$ among declared independent channels.
    \STATE $v_t^{c'} \leftarrow \mathrm{Evaluate}(x_t, \texttt{cross-channel}, c')$
    \IF{$|v_t^{c} - v_t^{c'}| > \epsilon_{\mathrm{ind}}$}
        \STATE $\beta_{c} \leftarrow \beta_{c} + 1$; $\beta_{c'} \leftarrow \beta_{c'} + 1$; \textbf{return} $\mathcal{P}_t$ \COMMENT{channels disagree: source unresolved}
    \ENDIF
    \STATE $\alpha_{c} \leftarrow \alpha_{c} + 1$; $\alpha_{c'} \leftarrow \alpha_{c'} + 1$; set state of $P$ to \textsc{Provisional}.
    \STATE \COMMENT{Stage 2b --- held-out generalization, the only admitting test}
    \STATE Draw $X_{\mathrm{heldout}} \sim \mathcal{D}_{\mathrm{heldout}}$ with $X_{\mathrm{heldout}} \cap \mathcal{H}_{\mathrm{search}} = \emptyset$, fixed before the run.
    \STATE $E(P) \leftarrow \{e = (x',y') : x' \in X_{\mathrm{heldout}},\ e.\mathrm{ind} = \textsc{true}\}$, with each stance set by the $\tau_{\mathrm{gen}}$ margin of §\ref{sec:methodology}.
    \IF{$\exists\, e \in E(P)$ with $\mathrm{contradict}(e,P) = 1$}
        \STATE Set state of $P$ to \textsc{Retracted}; recursively retract every principle listing $P$ as parent.
    \ELSIF{$|\{e \in E(P) : e.\mathrm{ind} \wedge e.\mathrm{heldout} \wedge \mathrm{support}(e,P)\}| \ge k$}
        \STATE Set state of $P$ to \textsc{Accepted}; $\mathcal{P}_{t+1} \leftarrow \mathcal{P}_t \cup \{P\}$.
    \ELSE
        \STATE Leave $P$ \textsc{Provisional} \COMMENT{support below $k$: search continues, state unchanged}
    \ENDIF
\ENDIF
\STATE \COMMENT{Standing re-verification: every new admissible record is also scored against each active}
\STATE \COMMENT{principle it bears on, so an \textsc{Accepted} principle is retracted by later contradiction,}
\STATE \COMMENT{not only at proposal time (cf.\ the mechanism-reversal trace, Appendix~\ref{app:protocol_reversal}).}
\FORALL{$P' \in \mathcal{P}_t$ with $\mathrm{state}(P') \neq \textsc{Retracted}$}
    \IF{a record $e$ admitted this round is linked to $P'$ and $\mathrm{contradict}(e,P') = 1$}
        \STATE Set state of $P'$ to \textsc{Retracted}; recursively retract every principle listing $P'$ as parent.
    \ENDIF
\ENDFOR
\STATE \textbf{return} $\mathcal{P}_{t+1}$.
\end{algorithmic}
\end{algorithm}

\section{Extended Benchmark Protocols and Failure Metric Definitions}
\label{app:protocol}

\paragraph{AutoResearchEval Benchmark.} AutoResearchEval~\citep{fei2026agents} evaluates multi-step targeted optimization across molecular properties, material bandgaps, and algorithmic code generation. Each harness is run across 5 independent seeds with an identical per-task monetary cap (\$2.00) and time budget.
\paragraph{Failure Metric Definitions.}
\begin{itemize}[nosep, leftmargin=12pt]
  \item \textbf{Official Score:} Raw target optimization metric reported by the host framework upon run termination.
  \item \textbf{Target Attainment:} Binary indicator indicating whether the best discovered candidate achieves $\ge 90\%$ of the task-specific global target optimum.
  \item \textbf{Verified Score:} Score evaluated independently by an external, sandboxed ground-truth testbed.
  \item \textbf{Valid Runs (\%):} Percentage of runs that complete all execution rounds without unhandled runtime crashes, tool timeouts, or syntax failure.
  \item \textbf{Critical Failure (\%):} Percentage of runs in which unverified, hallucinated tool outputs or corrupted feedback poisoned the agent's internal memory, leading to permanent loop collapse or degenerate repeating actions.
\end{itemize}

\end{document}